%% file: arxiv_main.tex
\documentclass[10pt,twocolumn,letterpaper]{article}

\usepackage[pagenumbers]{cvpr} % To force page numbers, e.g. for an arXiv version

\usepackage[accsupp]{axessibility}  % Improves PDF readability for those with disabilities.
\usepackage{multirow}
\usepackage{arydshln} % For the dashed line

\usepackage{listings}
\definecolor{cvprblue}{rgb}{0.21,0.49,0.74}
\usepackage[pagebackref,breaklinks,colorlinks,allcolors=cvprblue]{hyperref}

\def\paperID{31360} % *** Enter the Paper ID here
\def\confName{CVPR}
\def\confYear{2026}

\title{What and Where to Adapt: Structure–Semantics Co-Tuning for Machine\\Vision Compression via Synergistic Adapters}

\author{
	Shaobo Liu \quad Haobo Xiong \quad Kai Liu\thanks{Corresponding author: Kai Liu} \quad Yuna Lin \\
	Xidian University \\
	{\tt\small \{shaoboo.liu, hbxiong, ynling\}@stu.xidian.edu.cn} \quad {\tt\small kailiu@mail.xidian.edu.cn}
}

\begin{document}
\maketitle
\input{sec/0_abstract}    
\input{sec/1_introduction}
\input{sec/2_work}
\input{sec/3_method}
\input{sec/4_experiments}

\input{sec/5_conclusion}
{
    \small
    \bibliographystyle{ieeenat_fullname}
	\bibliography{all_right}
}

\input{sec/X_suppl}

\end{document}

%% file: sec/0_abstract.tex
\begin{abstract}
Parameter-efficient fine-tuning of pre-trained codecs is a promising direction in image compression for human and machine vision. While most existing works have primarily focused on tuning the feature structure within the encoder-decoder backbones, the adaptation of the statistical semantics within the entropy model has received limited attention despite its function of predicting the probability distribution of latent features. Our analysis reveals that naive adapter insertion into the entropy model can lead to suboptimal outcomes, underscoring that the effectiveness of adapter-based tuning depends critically on the coordination between adapter type and placement across the compression pipeline. Therefore, we introduce \textbf{S}tructure–\textbf{S}emantics \textbf{Co}-\textbf{T}uning (\textbf{S$^2$-CoT}), a novel framework that achieves this coordination via two specialized, synergistic adapters: the Structural Fidelity Adapter (SFA) and the Semantic Context Adapter (SCA). SFA is integrated into the encoder-decoder to preserve high-fidelity representations by dynamically fusing spatial and frequency information; meanwhile, the SCA adapts the entropy model to align with SFA-tuned features by refining the channel context for more efficient statistical coding. Through joint optimization, S$^2$-CoT turns potential performance degradation into synergistic gains, achieving state-of-the-art results across four diverse base codecs with only a small fraction of trainable parameters, closely matching full fine-tuning performance. Code is available at \url{https://github.com/Brock-bit4/S2-CoT}.
\end{abstract}

%% file: sec/1_introduction.tex
\section{Introduction}
\label{sec:introduction}

\begin{figure}[t]
	\centering
	\includegraphics[width=1.0\columnwidth]{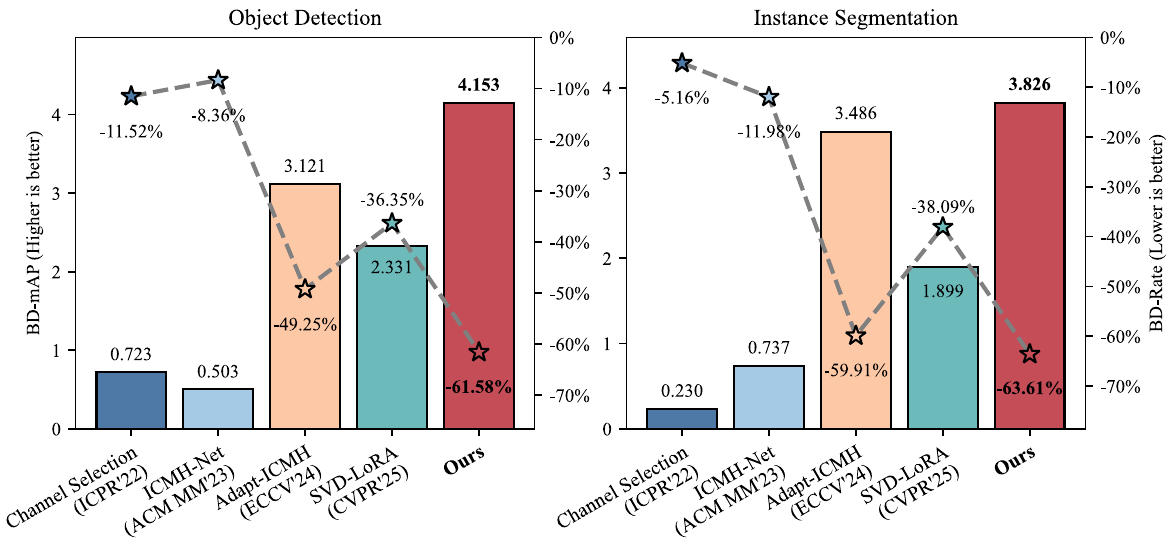} % Reduce the figure size so that it is slightly narrower than the column. Don't use precise values for figure width.This setup will avoid overfull boxes.
	\caption{Rate-accuracy performance comparison of PEFT methods on the COCO2017 using the  $ \textit{\shortstack{Cheng2020-anchor}} $ base codec. Our method sets a new state-of-the-art by simultaneously achieving the highest accuracy gains (BD-mAP$\uparrow$) and the largest bitrate savings (BD-Rate$\downarrow$) across both object detection and instance segmentation tasks, significantly outperforming prior works.}
	\label{fig1_ours_best_paper}
\end{figure}
The proliferation of large-scale AI \cite{He2016, Vaswani2017, Liu2021} necessitates efficient Image Compression for Machine and Human vision (ICMH) \cite{Bai2022, Choi2022, Codevilla2021, Yang2021}. This critical research area aims to produce a unified low-bitrate bitstream that supports high visual fidelity for humans and robust performance for downstream machine tasks like object detection and instance segmentation \cite{Chen2023, Feng2023, Fischer2025, Guo2024, Liu2022, Liu2023, Liu2023a, Liu2024, Xue2024, Yang2021, Yin2025, Zhang2024, Zhao2025}. A prevalent trend is to adapt pre-trained, human-oriented base codecs using Parameter-Efficient Fine-Tuning (PEFT) \cite{He2022, Li2024}, avoiding the prohibitive training cost and storage overhead of training task-specific models from scratch.

The success of PEFT in Learned Image Compression (LIC) hinges on strategically inserting trainable and lightweight adapters within the compression pipeline \cite{Shen2023, Chen2023, Feng2023a, Li2024}. This pipeline consists of two core components: the encoder-decoder for feature structure and the entropy model for statistical semantics modeling. However, existing methods exhibit a pronounced imbalance: prior works \cite{Wang2021, Choi2022, Zhao2025} focus almost exclusively on adapting the feature structure within the backbone, while the entropy model’s statistical semantics have been largely neglected \cite{Minnen2020, Iliopoulou2025}.

Moreover, the sensitivity of the entropy model renders naive adapter transplantation ineffective. As confirmed in \cref{tab1_sfma}, a structural adapter from \cite{Li2024} shows a stark performance collapse when applied to the entropy model, while stacking it across both modules yields negligible gains at the double cost of parameter overhead.

We argue that it does not stem from inability to adapt the entropy model, but from a fundamental mismatch in adapter design. An adapter specialized for structural representation in the encoder-decoder is architecturally and functionally ill-suited for the statistical semantic modeling required by the entropy model. This motivates a principled investigation into \textbf{what types of adapters are effective and where they should be placed across the compression pipeline}. 
\begin{table}[t]
	\caption{A demonstration of adapter-architecture mismatch. The structure-focused adapter from \cite{Li2024} proves ineffective when applied solely to the entropy model ($h_a$, $h_s$) and yields no discernible benefit when combined with the encoder-decoder ($g_a$, $g_s$). This highlights the need for targeted adapter design over naive reuse.}
	\label{tab1_sfma}
	\centering
	\small
	\setlength{\tabcolsep}{1.0mm}
	\begin{tabular}{ccccc}
		\toprule
		\multirow{2}{*}{\shortstack{Adapter\\Placement}} & \multirow{2}{*}{Adapter Type} & \multicolumn{2}{c}{Classification} & \multirow{2}{*}{\shortstack{Params\\$\downarrow$(M)}} \\
		& & BD-Rate$\downarrow$ & BD-Acc$\uparrow$ & \\
		\midrule
		$g_a$, $g_s$ & \textbf{Structural} & -82.00\% & 18.71 & 0.28 \\
		$h_a$, $h_s$ & \textbf{Structural} & -1.56\% & 0.27 & 0.26 \\
		combined & \textbf{Structural} & -81.27\% & 18.80 & 0.55 \\
		\bottomrule
	\end{tabular}
\end{table}

Our work answers this question by proposing \textbf{S}tructure-\textbf{S}emantics \textbf{C}o-\textbf{T}uning (\textbf{S$^2$-CoT}), a novel framework comprising two specialized, synergistic adapters that exploit the distinct architectural properties of the compression pipeline. The encoder-decoder backbone, which operates on spatially expansive feature representations, is augmented with a Structural Fidelity Adapter (SFA) designed to preserve structural priors by dynamically fusing spatial and frequency information \cite{Fu2024, Li2024, Li2024a}. Conversely, the entropy model utilizes a compact latent space where channel-wise statistical redundancies are dominant. We therefore propose a lightweight Semantic Context Adapter (SCA) to explicitly model inter-channel dependencies, also inspired by recent advances in channel-wise entropy modeling \cite{Minnen2020, Iliopoulou2025}. This co-tuning of SFA and SCA unlocks a powerful synergy, as shown in \cref{fig1_ours_best_paper}. We summarize the main contributions as:
\begin{itemize}
	\item We theoretically and experimentally validate the structure–semantics synergy in ICMH, showing that the isolated adaptation of either the backbone or the entropy model inevitably leads to suboptimal performance.
	\item We identify a key insight that adapter effectiveness depends critically on matching type (\eg, structural vs. semantic) to its placement in the compression pipeline.
	\item We propose a highly effective S$^2$-CoT implementation via two specialized, synergistic adapters: SFA and SCA. SFA, integrated into the encoder-decoder, preserves high-fidelity representations by dynamically fusing spatial and frequency information. As the first PEFT module validated for the entropy model, SCA adapts the entropy model to align with SFA-tuned features, refining channel context for more efficient statistical coding.
	\item Our unified S$^2$-CoT framework achieves state-of-the-art (SOTA) results across various base codecs and entropy models, rivaling the performance of full fine-tuning with only a fraction of the trainable parameters.
\end{itemize}

%% file: sec/2_work.tex
\section{Related Work}
\label{sec:work}

\subsection{Image Compression for Machine Vision}
Learned image compression backbones, evolving from CNN-based to Transformer-based, perform exceptionally well for human perception. However, these backbones do not directly translate to machine vision tasks \cite{Codevilla2021, Liu2021a, Bai2022, Chen2023, Xue2024, Zhang2024, Liu2024, Yin2025}. Furthermore, training full fine-tuning models from scratch for each task is computationally prohibitive and yields inflexible models.

Consequently, some works explored complex unified frameworks \cite{Zhang2024, Guo2024, Yin2025} and even large language model applications \cite{Murai2024, Xue2024, Song2025}. For instance, Zhang \etal \cite{Zhang2024} proposed an integrated framework with multi-path aggregation for a unified representation, while Guo \etal \cite{Guo2024} developed a multi-task co-embedding approach to generalize to unforeseen tasks. Despite their progress, these methods often require intricate, multi-branch architectures or multiple bitstreams, bringing substantial overhead. Our work differs by utilizing the lightweight adapter technique that can be applied to the majority of CNN-based and Transformer-based LIC models, enabling high-performance task support.

\subsection{Parameter-Efficient Adaptation in ICMH}
PEFT, often realized through lightweight techniques like prompts and adapters \cite{Chen2022, Chen2023a, Tsubota2023, Li2024, Park2025}, has recently been introduced to ICMH to adapt pre-trained base codecs with minimal overhead \cite{Shen2023}. TransTIC \cite{Chen2023}, for instance, successfully applied prompt-based tuning by inserting instance-specific and task-specific cues into Transformer-based codecs. Other works, like Adapt-ICMH \cite{Li2024}, have developed lightweight adapters by modulating spatial and frequency domains, while lacking a mechanism for their dynamic, adaptive fusion. Furthermore, a common thread in these methods is their exclusive focus on modulating the feature flow within the main encoder-decoder backbone. In contrast, to the best of our knowledge, our work is the first to introduce targeted adaptation into the entropy model and, more critically, the first to systematically study the synergistic interplay between structural adaptation (via SFA) and statistical model adaptation (via SCA).

\subsection{Entropy Model Optimization}
The entropy model is the statistical core of a learned codec, responsible for accurately estimating the probability distribution of the latent features. Its design has progressed from simple factorized priors \cite{Balle2017} to more powerful hyperprior \cite{Balle2018} and autoregressive models \cite{Minnen2018, Minnen2020}, such as channel-wise autoregressive schemes, that capture complex dependencies in the latent space. However, prior works \cite{Lee2019} have noted that the static models are incapable of capturing input-specific variations \cite{Choi2022}, and that neglecting channel interactions is particularly detrimental for dense prediction tasks \cite{Li2024, Wang2021}. To address these specific limitations, the SCA is designed as a lightweight plug-in that integrates into and enhances the entropy model's existing channel-wise structure. Rather than replacing the entire architecture, it refines the contextual information used for probability prediction. Crucially, this modular design allows the SCA to be applied to various types of entropy models, from hyperprior-based to autoregressive-based architectures.

%% file: sec/3_method.tex
\section{Proposed Method}
\label{sec:method}

\begin{figure*}[!htbp]
	\centering
	\includegraphics[width=2\columnwidth]{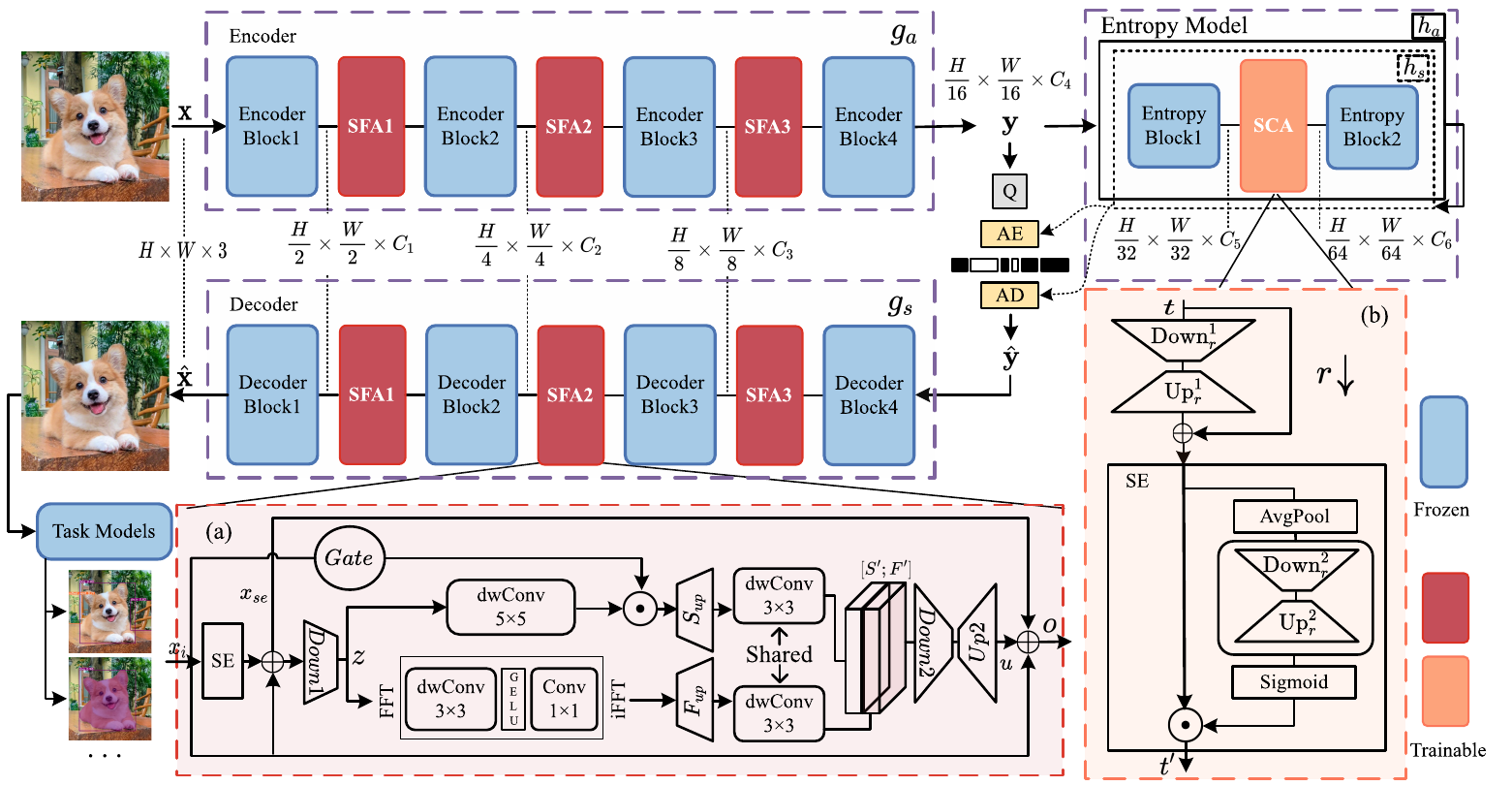} % Reduce the figure size so that it is slightly narrower than the column. Don't use precise values for figure width.This setup will avoid overfull boxes.
	\caption{Overview of our proposed S$^2$-CoT. It includes the Encoder, Decoder, Entropy Model, and Task Models. The colors of the modules indicate their training status: the base codecs and downstream task models are frozen (indicated in blue), while the trainable SFA and SCA are in red and orange, respectively. Here, $ \odot $ denotes the Hadamard product and $\oplus$ denotes element-wise addition.}
	\label{fig2_overview}
\end{figure*}
\subsection{Synergy Motivation and Theoretical Basis}
\label{sec:mismatch_proof}

We utilize a stage-wise freezing experiment to demonstrate that this isolated structural tuning breaks the essential statistical co-dependency between the encoder-decoder and entropy model, establishing the need for co-tuning.
\begin{table}[!htbp]
	\caption{Analysis of a stage-wise freezing experiment on the $ \textit{Lu2022-TIC} $ \cite{Lu2021} for object detection on the COCO2017 dataset. In this experiment, specific network components were frozen to show the effects of fine-tuning only the encoder-decoder ($g_a$, $g_s$) or fine-tuning them along with the entropy model ($h_a, h_s$).}
	\centering
	\small
	\begin{tabular}{ccccccccc}
		\toprule
		$g_a$    & $g_s$    & $h_a$    & $h_s$    & bpp $\downarrow$   & mAP $\uparrow$   & $\rho \downarrow$ \\
		\midrule
		0-7   & 0-7   & 0-3   & 0-3   & 0.0639 & 37.63& 0.239 \\
		1,3,5   & 2,4,6   & \textbackslash{} & \textbackslash{} & 0.0829 & 37.21&0.289 \\
		0-7   & 0-7   & \textbackslash{} & \textbackslash{} & 0.0881 & 37.32& 0.314 \\
		1,3,5   & 2,4,6   & 0-3   & 0-3   & 0.0685 & 37.42&  0.260 \\
		\bottomrule
	\end{tabular}
	\label{analysis1}
\end{table}

As shown in \cref{analysis1}, adapting only the encoder-decoder (row 3) yields a substantial bitrate increase and mAP drop versus full fine-tuning (row 1), supported by a sharp rise in normalized latents' correlation $\rho$ \cite{Zhu2022}. This failure demonstrates that structural adaptation in the absence of simultaneous semantic adaptation is fundamentally unsound.

We attribute this degradation to the latent distribution deviation induced by isolated structural tuning in the encoder-decoder stage. When the encoder-decoder ($g_a, g_s$) is tuned, the resulting feature distribution ($y'$) shifts statistically, causing the frozen entropy model to fail in accurate modeling and yield suboptimal parameters $(\mu', \sigma')$. This statistical breakdown introduces a measurable bitrate penalty ($\Delta R$). The actual compressed rate $R(\hat{y})$ is decomposed into the ideal rate plus the penalty term, demonstrating the quantitative consequence of the statistical shift in \cref{eq_s1}:
\begin{align}
	R(\hat{y}) &= \mathbb{E} \left[ -\log_2 (p_{\hat{y}|\hat{z}} (\hat{y} \mid \hat{z})) \right] \\
	&= \mathbb{E} \left[ -\log_2 (p_{\hat{y}|\tilde{z}} (\hat{y} \mid \tilde{z})) \right] + \Delta{R}
	\label{eq_s1}
\end{align}

%Detailed formula derivation is provided in appendix.
Detailed formula derivation and theoretical analysis are provided in \cref{sec:supTheoreticalAnalysis} \& \cref{sec:supgeneralizationSynergy} for clarity and completeness.

\subsection{S$^2$-CoT Design and Implementation}
Minimizing this $\Delta R$ and restoring optimal coding efficiency mandates a structural-semantic co-tuning approach. Our S$^2$-CoT framework achieves this via the synergy of SFA and SCA, as shown in \cref{fig2_overview}. The SFA provides the necessary optimized structural features. The SCA then functions as the statistical corrective mechanism, adapting the entropy model to update the suboptimal parameters $(\mu', \sigma')$ towards the ideal set. The following description is based on its implementation within $ \textit{Lu2022-TIC} $ \cite{Lu2021}.

Specifically, the design of both adapters stems from the distinct roles of the two main pipeline components. Since the encoder-decoder operates on spatially-structured feature maps where structural and textural information is paramount, the SFA is designed to directly modulate both spatial and frequency domains, enhancing the codec's ability to preserve structural fidelity. In contrast, the entropy model operates on a more abstract, channel-wise statistical representation. Consequently, the SCA is designed to focus on channel interactions and semantic context to improve the efficiency of the latent distribution modeling.

Architecturally, the SFA modules are strategically interleaved at multiple stages within the top-3 encoder $ g_a $ and bottom-3 decoder $ g_s $. This placement allows for multi-resolution adaptation of the feature maps. During adaptation, the parameters of the base codecs remain frozen, and only the proposed SFA and SCA adapters are jointly optimized. For a given input image $\mathbf{x}$, the adapters are trained by minimizing the following loss function:
\begin{equation}
	\mathcal{L} = \mathcal{R}(\hat{\mathbf{y}}) + \lambda \cdot \mathcal{D}(\mathbf{x}, \hat{\mathbf{x}}; \mathcal{G})
	\label{loss_function}
\end{equation}
where  $ \mathcal{R} $ is the estimated bitrate of the quantized latent representation $ \hat{\mathbf{y}} $. The term $ \mathcal{D} $ represents a task-specific perceptual distortion, computed using a frozen, pre-trained model $ \mathcal{G} $ that extracts features from both the original image $\mathbf{x}$ and the reconstructed image $\hat{\mathbf{x}}$. The hyperparameter $\lambda$ controls the trade-off between rate and task-specific distortion.

\subsection{Structural Fidelity Adapter}
The SFA is a compact module specifically designed to enhance the structural fidelity within the encoder-decoder, as illustrated in \cref{fig2_overview}(a). The SFA fundamentally adopts a novel dual-branch structure for spatial-frequency modulation, which ultimately converges in a soft fusion mechanism. Specifically, our design seamlessly integrates: (I) Channel Excitation and Bottleneck Projection, (II) Spatial-Frequency Dual-Branch Modulation, and (III) Soft Fusion.

\paragraph{Stage I: Channel Excitation and Bottleneck Projection.} For each feature map \( x_i \in \mathbb{R}^{C_i \times \frac{H}{2^i} \times \frac{W}{2^i}} \) at stage \( i \in \{1, 2, 3\} \), we first employ a Squeeze-and-Excitation(SE) Attention block \cite{Hu2018} to recalibrate channels. This initial step allows the adapter to focus on the most informative channels. The enhanced feature map $x_{se}$ is combined with the original input $ x_i $ and projected via a $1\times1$ convolutional bottleneck to reduce dimensionality, as formulated in \cref{bottleneck_proj}:
\begin{equation}
	z = Down1(x_i \oplus x_{se}), \quad z \in \mathbb{R}^{C' \times \frac{H}{2^i} \times \frac{W}{2^i}}, \quad C' \ll C_i
	\label{bottleneck_proj}
\end{equation}
Here, $ C' $ is the reduced channel dimension. This projection $ Down1 $ reduces computational complexity and generates a unified, low-rank representation $z$. 

\paragraph{Stage II: Spatial-Frequency Dual-Branch Modulation.}
The representation $ z $ is processed in two parallel branches to independently capture distinct feature properties.

\textit{(i) Spatial Branch.}
The spatial branch is designed to model local structural patterns. It employs a depth-wise convolution to capture fine-grained spatial interactions, which are then modulated by a $ 1\times1 $ convolutional gate that takes the original, high-dimensional feature $ x_i $ as input. The resulting spatial modulation $ S $ is formulated as:
\begin{equation}
	\begin{aligned}
		S_{up} &= dwConv_{5 \times 5}(z) \odot Gate(x_i), \\
		S &=  W_{su} \cdot \delta(S_{up}).
	\end{aligned}
	\label{spatial_branch}
\end{equation}
where $ \delta $ is the ReLU activation, and $ W_{\text{su}} $ is an up-projection convolution that restores the channel dimension to $ C_i $.

%\textit{(ii) Frequency Branch.} The frequency branch $F$ modulates textural information using FFT/IFFT. As defined in Eq. \cref{freq_branch}, it operates on the amplitude spectrum ($\mathcal{A}$) and phase spectrum ($\mathcal{P}(z)$), applying an adaptive amplitude modulation ($\sigma(X)$) to precisely control texture features. Detailed derivation and architectural analysis are provided in appendix.

\textit{(ii) Frequency Branch.} 
The frequency branch operates on the frequency representation of the feature to modulate textural information. We transform $z$ to the frequency domain using Fast Fourier Transform (FFT). The complete frequency modulation $ F $ is defined as \cref{freq_branch}:
\begin{equation}
	\begin{aligned}
		% Define an intermediate variable for clarity and brevity
		\hat{z} &= \mathcal{F}(z),\mathcal{A} = |\hat{z}|, \mathcal{P} = \angle(\hat{z}), \\
		X &= W_f \cdot \phi\left( dwConv_{3 \times 3}(\mathcal{A}) \right), \\
		F_{up} &= \mathcal{F}^{-1} \left( X \cdot \sigma \left( X \right) \cdot e^{j \mathcal{P}} \right), \\
		F &= W_{fu} \cdot \delta(F_{up}).
	\end{aligned}
	\label{freq_branch}
\end{equation}
where $\phi(\cdot)$ and $\sigma(\cdot)$ denote the GELU and Sigmoid, while $\mathcal{F}(\cdot)$ and $\mathcal{F}^{-1}(\cdot)$ denote FFT and inverse FFT, respectively. Applying FFT to $z$ yields the amplitude spectrum $\mathcal{A}$ and phase spectrum $\mathcal{P}$. Both $ W_f $ and $  W_{fu} $ are $1 \times 1$ convolutions but serve different purposes: $ W_f $ modulates amplitude in the frequency domain, while $ W_{fu} $ expands channel dimensionality and refines features after inverse transformation. Phase term $ e^{j \mathcal{P}} $ is retained to stabilize training.
%Although $ e^{j \mathcal{P}(z)} $ carries the phase information, we avoid learning directly from phase due to its instability during training.

%where $\mathcal{F}(\cdot)$ and $\mathcal{F}^{-1}(\cdot)$ denote FFT and inverse FFT, respectively; $\mathcal{A}$ and $\mathcal{P}$ denote the amplitude and phase spectra; $\phi(\cdot)$, $\sigma(\cdot)$, and $\delta(\cdot)$ denote GELU, Sigmoid, and ReLU, respectively. $W_f$ and $W_{fu}$ are $1 \times 1$ convolutions for amplitude refinement and output projection.

\begin{figure*}[!htbp]
	\centering
	\begin{tabular}{cc:c}
		% Row 1: Three images
		\includegraphics[width=0.31\textwidth]{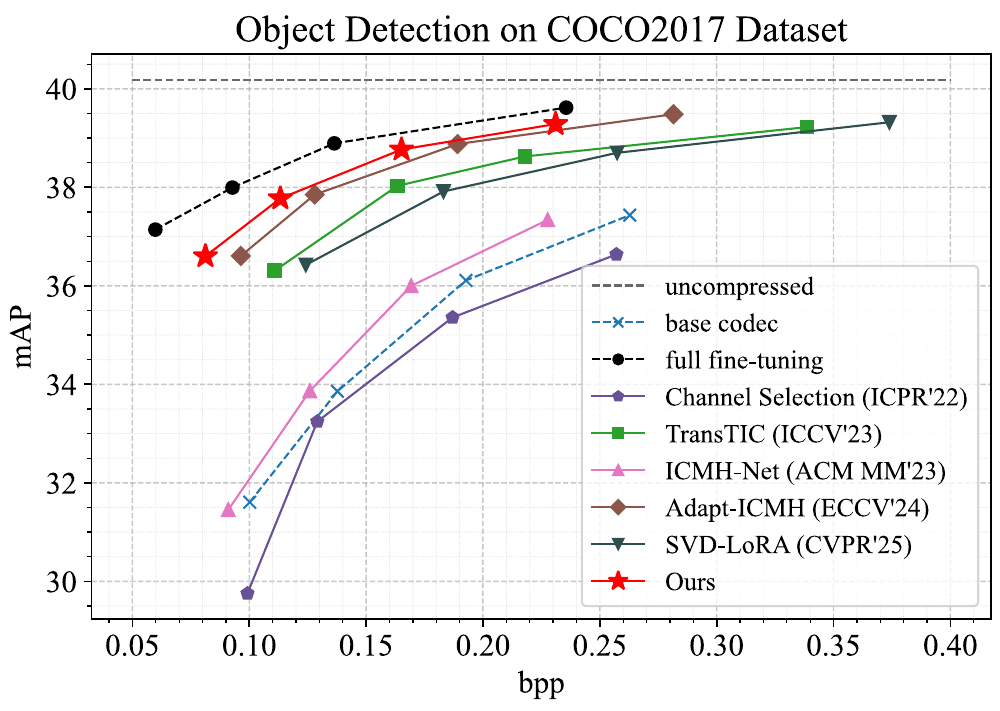} &
		\includegraphics[width=0.31\textwidth]{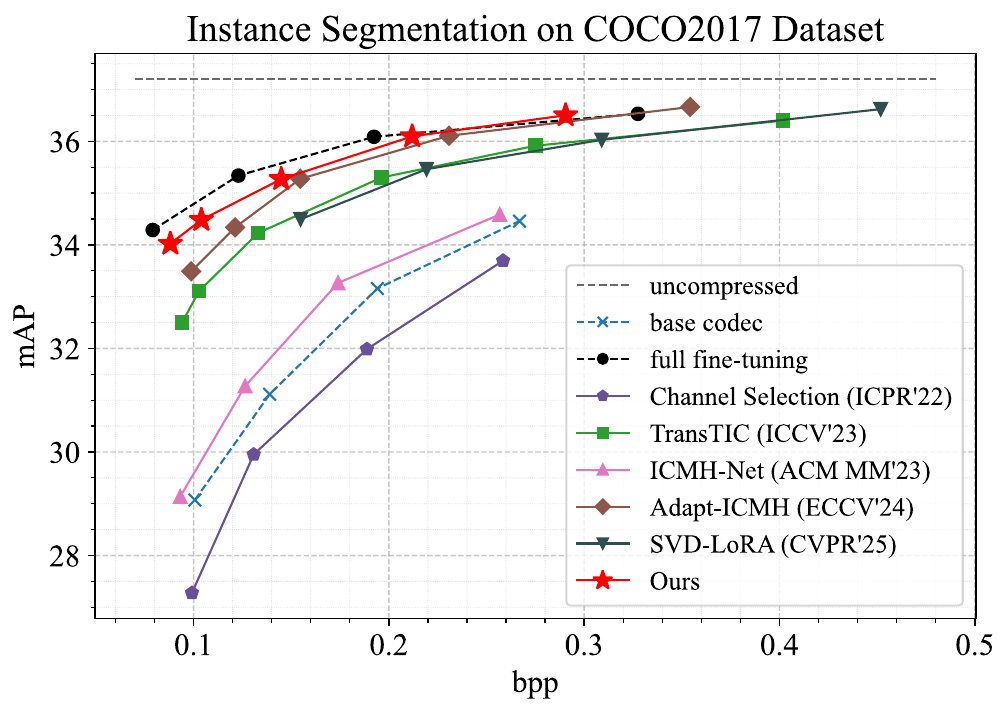} &
		\includegraphics[width=0.31\textwidth]{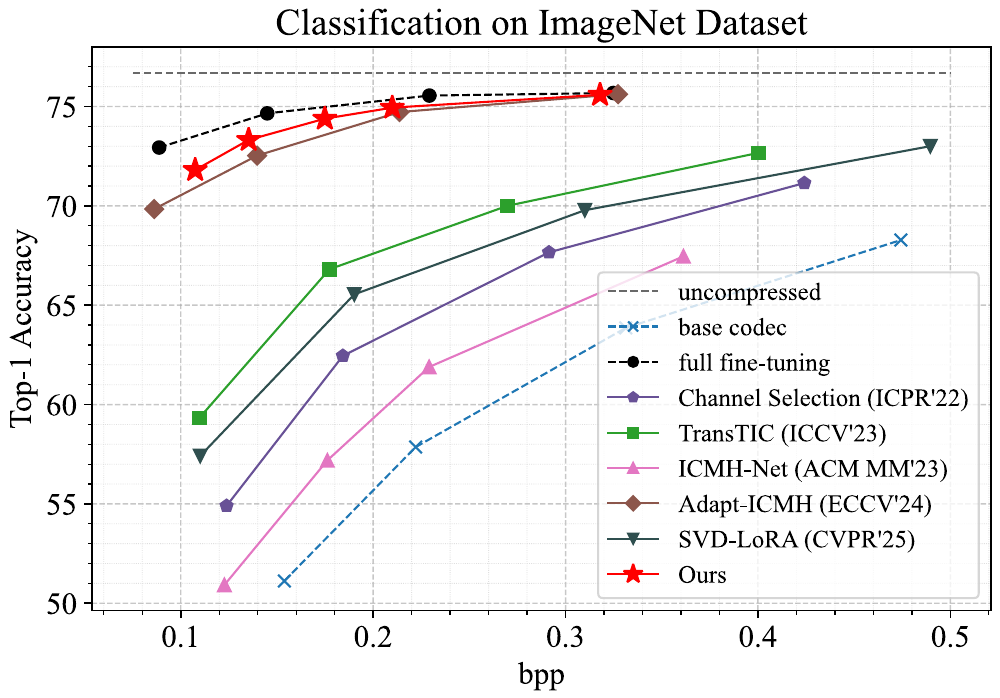} \\
		% Row 2: Empty merged cell, and the (b) label for the image above
		\multicolumn{2}{c}{} & (b) \\
		% Row 3: Three images
		\includegraphics[width=0.31\textwidth]{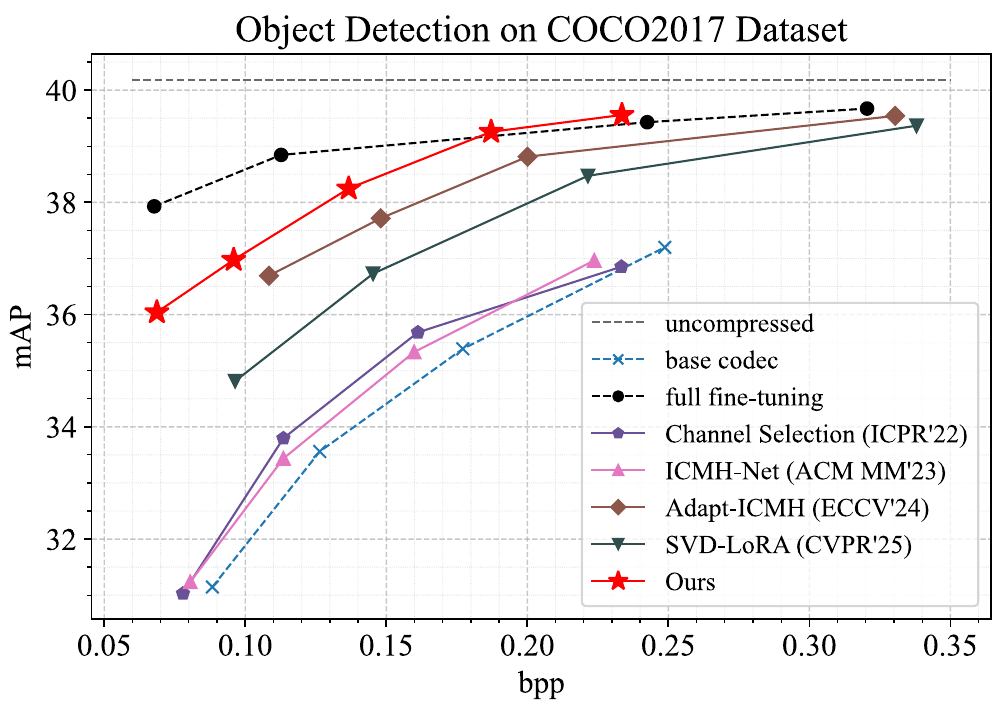} &
		\includegraphics[width=0.31\textwidth]{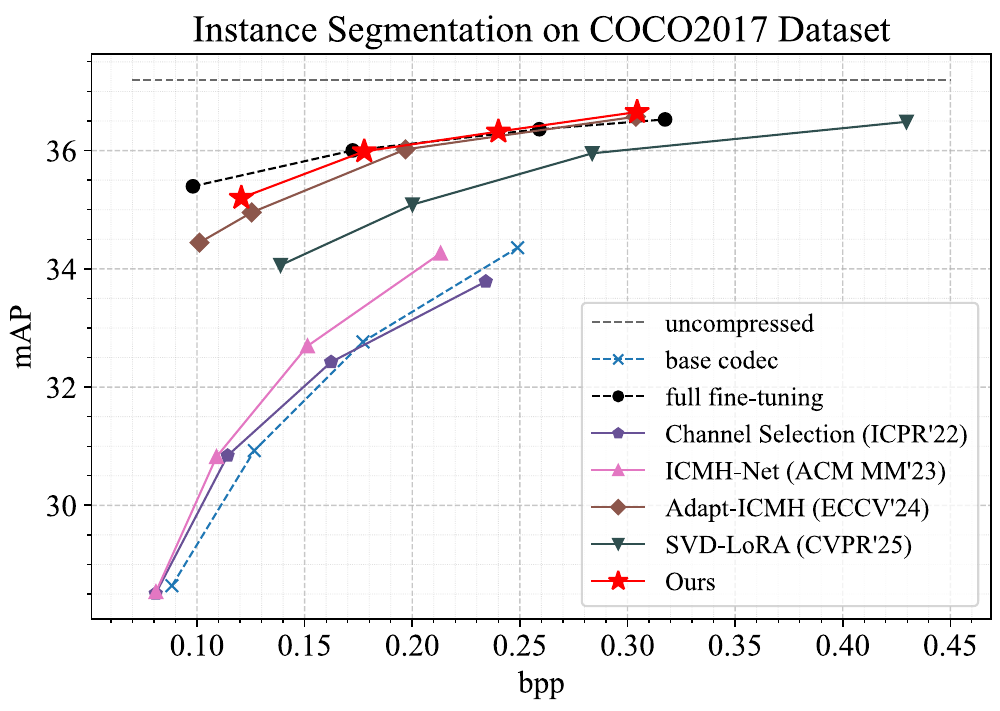} &
		\includegraphics[width=0.31\textwidth]{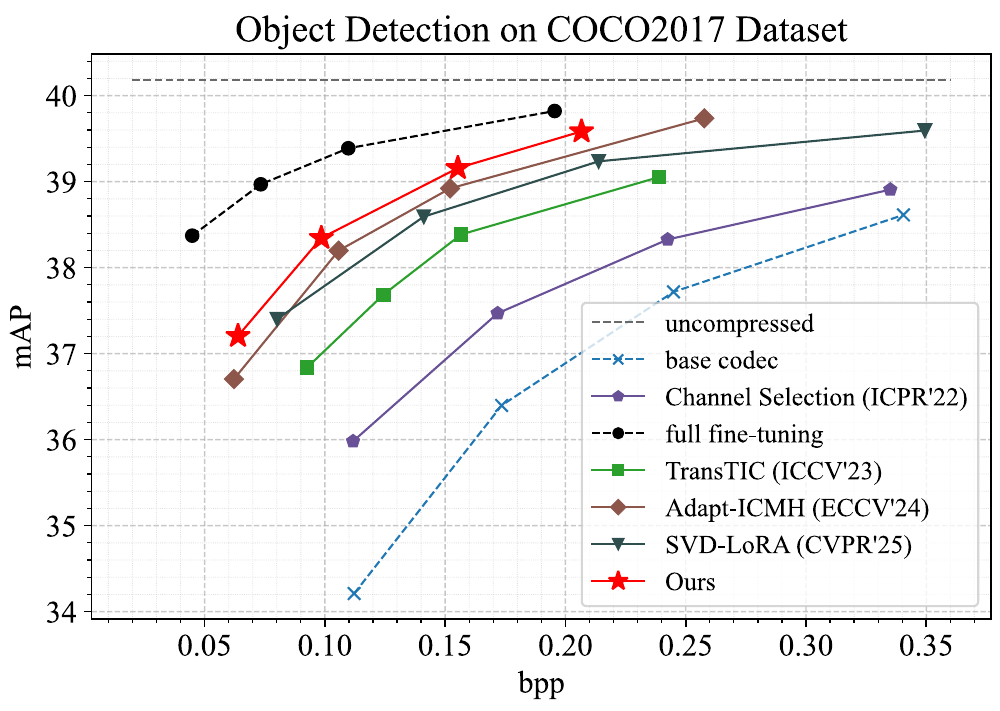} \\
		% Row 4: Merged cell for (a) label, and the (c) label
		\multicolumn{2}{c}{(a)} & (c) \\
	\end{tabular}
	\caption{Comparison of rate-accuracy performance across various tasks and base codecs. (a) Object detection and instance segmentation results on the $ \textit{Lu2022-TIC} $ (top row) and $ \textit{\shortstack{Cheng2020-anchor}} $ (bottom row) base codecs. (b) Classification performance on the $ \textit{Lu2022-TIC} $ base codec. (c) Object detection performance on the $ \textit{DCAE} $ base codec. Results for the fourth base codec $ \textit{ELIC} $ are in \cref{sec:supMoreResultsOnTheDiverseBaseCodec}.}
	\label{main_comparison}
\end{figure*}

\paragraph{Stage III: Soft Fusion.}
Finally, we introduce a soft fusion stage to learn the optimal, non-linear integration of the two branches. This stage first pre-processes the spatial modulation $ S $ and frequency modulation $ F $ using a shared depth-wise convolution ($ \mathrm{dwConv}^{\text{shared}}_{3\times3}(\cdot) $) to generate intermediate features $ S' $ and $ F' $. These are then further fused along the channel dimension for joint representation learning. To enable inter-branch interaction with minimal overhead, $u$ is processed through a lightweight channel mixing module consisting of two consecutive $1 \times 1$ convolutions, referred to as $Down2$ and $Up2$. Specifically, $Down2$ compresses the channel dimension from $2C_i$ to half of $C_i$, and $Up2$ expands it back to $C_i$, facilitating residual connection with $x_i$ and $x_{se}$. The final fused output $ o $ is computed as \cref{fusion_block}:

\begin{equation}
	\begin{aligned}
		S'&=\mathrm{dwConv}^{\text{shared}}_{3\times3}(S), F'=\mathrm{dwConv}^{\text{shared}}_{3\times3}(F), \\
		u &= Up2 \cdot \delta\left( Down2 ([S';F']) \right), \\
		o &= x_i \oplus x_{se} \oplus u.
	\end{aligned}
	\label{fusion_block}
\end{equation}

\subsection{Semantic Context Adapter}
In learned image compression, the entropy model is pivotal for bitrate control via probability estimation. A key characteristic of its operational domain is the severe spatial downsampling of its input features (e.g., latent representation $ \mathbf{y} $ and hyperprior feature $ t $ ), often by a factor of 64$ \times $ relative to the original image $ \mathbf{x} $. This aggressive reduction renders spatial modeling inherently sparse and less reliable. Conversely, the channel dimension becomes the dominant axis for information, making it the most principled and effective domain for fine-grained, task-specific adaptation.

Motivated by this, we propose the SCA, a lightweight adapter designed to enhance the statistical context within the entropy model, as illustrated in \cref{fig2_overview}(b). First, the SCA performs feature refinement via a residual bottleneck projection, which enriches the feature representation of each channel. Concurrently, it performs feature recalibration using a Squeeze-and-Excitation block that computes a multiplicative, channel-wise attention vector to re-weight feature importance. The operation is formulated as \cref{sca}:  
\begin{equation}
	t' = 
	\left( t + \text{Up}_{r}^1 \left( \text{Down}_{r}^1(t) \right) \right)
	\cdot 
	\underbrace{\sigma \left( \text{Up}_{r}^2 \left( \text{Down}_{r}^2(P(t)) \right) \right)}_{\scriptsize \text{SE Attention}}
	\label{sca}
\end{equation}
where $ P(\cdot)$ denotes the global average pooling operator, $ \sigma $ is the Sigmoid function, and $r$ is the channel reduction ratio. The terms $\text{Down}_{r}^1, \text{Up}_{r}^1$ and $\text{Down}_{r}^2, \text{Up}_{r}^2 $ represent the down-projection and up-projection $1\times1$ convolutions for the bottleneck projections and SE Attention, respectively.

\begin{table*}[!htbp]
	\caption{Quantitative comparison for object detection and instance segmentation tasks. Our S$^2$-CoT nears full fine-tuning and outperforms other PEFT methods. Best results are in \textbf{bold}, second-best are \underline{underlined}. See \cref{sec:supMoreMainExperimentalResults} for full $ \textit{Cheng2020-anchor} $-based results.}
	\label{tic_results}
	\centering
	\small
	\begin{tabular}{cccccccc}
		\toprule
		\multirow[c]{2}{*}{Base} & \multirow[c]{2}{*}{Method} & \multirow[c]{2}{*}{Venue} & \multicolumn{2}{c}{Object Detection} & \multicolumn{2}{c}{Instance Segmentation} & \multirow[c]{2}{*}{\shortstack{Trainable\\Params$\downarrow$ (M)}} \\
		%		\cline{4-7}
		& & & BD-Rate$\downarrow$ & BD-mAP$\uparrow$ & BD-Rate$\downarrow$ & BD-mAP$\uparrow$ & \\
		\midrule
		\multirow[c]{6}{*}{$ \textit{Lu2022-TIC} $}
		& full fine-tuning & -- & -73.943\% & 4.511 & -67.977\% & 3.755 & 7.51 (100.00\%) \\ %\cline{2-8}
		& Channel Selection & ICPR\textquotesingle22 & 6.849\% & -0.550 & 16.511\% & -0.949 & 0.92 (12.25\%) \\
		& TransTIC & ICCV\textquotesingle23 & -46.301\% & 2.768 & -46.521\% & 2.690 & 1.62 (21.57\%) \\
		& ICMH-Net & ACM MM\textquotesingle23 & -9.080\% & 0.625 & -10.772\% & 0.654 & 3.98 (53.00\%) \\
		& Adapt-ICMH & ECCV\textquotesingle24 & -55.150\% & 3.547 & -52.407\% & 3.208 & 0.29 (3.86\%) \\
		& SVD-LoRA & CVPR\textquotesingle25 & -39.927\% & 2.207 & -42.431\% & 1.938 & \textbf{0.09} (1.20\%) \\
		& \underline{Ours ($C'$=32, $r$=32)} & -- & \underline{-60.276\%} & \underline{3.852} & \underline{-60.255\%} & \underline{3.386} & \underline{0.28} (3.73\%) \\
		& \textbf{Ours} & -- & \textbf{-60.824\%} & \textbf{4.014} & \textbf{-61.784\%} & \textbf{3.480} & 0.42 (5.59\%) \\
		\midrule
		\multirow[c]{2}{*}{ $ \textit{\shortstack{Cheng2020\\-anchor}} $ }
		& full fine-tuning & -- & -59.015\% & 4.699 & -74.052\% & 3.869 & 26.60 (100.00\%) \\ 
		& \textbf{Ours} & -- & \textbf{-61.578\%} & \textbf{4.153} & \textbf{-63.607\%} & \textbf{3.826} & 0.74 (2.78\%) \\
		\bottomrule
	\end{tabular}
\end{table*}

%% file: sec/4_experiments.tex
\section{Experiments}
\label{sec:experiments}

\subsection{Experimental Settings and Datasets}
To verify the versatility of our method, we conducted primary experiments on two distinct, mainstream codecs: the Transformer-based codec $ \textit{Lu2022-TIC} $ \cite{Lu2021}, which utilizes a hyperprior entropy model, and the CNN-based codec $ \textit{Cheng2020-anchor} $ \cite{Cheng2020} featuring a spatial autoregressive entropy model. For all experiments, the base codec parameters are frozen, loaded from the CompressAI \cite{Begaint2020} library.

We use the COCO2017 \cite{Lin2014} training set for training and its validation set for testing. The Kodak dataset is additionally used as a validation set. These datasets are widely recognized as standard benchmarks for image compression and downstream computer vision tasks, facilitating direct comparability with existing SOTA methods. All images are resized and randomly cropped to 256×256 patches during training. The loss is computed using frozen, pre-trained task-specific models: Faster R-CNN \cite{Ren2015} for detection and Mask R-CNN \cite{He2017} for segmentation from the Detectron2 library. We adopt the ImageNet dataset \cite{Deng2009} when extending our method to classification tasks. Further implementation details, including the experimental equipment and hyperparameter settings, are provided in \cref{sec:supTaskPerceptualDistortionLoss} and \cref{sec:supExperimentalSettingsOfTraining}.

\subsection{Evaluation Metrics}
Following the evaluation setting of \cite{Chen2023}, we quantify the compression rate in bits per pixel (bpp). Task performance is measured via mean Average Precision (mAP) for detection and segmentation, and Top-1 for classification.

The overall rate-performance trade-off is assessed using the Bjøntegaard Delta (BD) methodology. BD-Rate is reported to quantify the average bitrate savings at an equal level of task accuracy. For detection and segmentation, we utilize BD-mAP, which extends the traditional formulation by replacing PSNR with mAP to measure average accuracy improvements at equivalent bitrates.

\subsection{Main Results and State-of-the-Art Comparison}
As visualized in the rate-accuracy curves in \cref{main_comparison}, the S²-CoT framework consistently achieves an excellent trade-off compared to all competing PEFT methods (including Channel Selection \cite{Liu2022}, TransTIC \cite{Chen2023}, ICMH-Net \cite{Liu2023a}, Adapt-ICMH \cite{Li2024} and SVD-LoRA \cite{Park2025}) across both detection and segmentation. It can be noticed that S$^2$-CoT performance closely approaches the upper bound set by full fine-tuning, an achievement that prior works fail to attain.

The quantitative results presented in \cref{tic_results} substantiate this advantage, demonstrating superior performance through both BD-rate savings and BD-mAP gains. Remarkably, even the most parameter-constrained model remains highly competitive, underscoring the exceptional efficiency of the S$^2$-CoT. An extended experiment on a more recent CNN-based base codec $ \textit{ELIC} $ \cite{He2022a} also achieves SOTA performance, with the corresponding results presented in Appendix K. Collectively, these results validate that our coordinated adaptation of both codec and entropy model is a more effective and efficient paradigm for LIC.

\begin{table*}[!htbp]
	\caption{Ablation study on different components of SFA.}
	\label{sfa_ablation}
	\centering
	\small
	\begin{tabular}{ccccccc}
		\toprule
		\multirow{2}{*}{Stage} & \multirow{2}{*}{Module} & \multicolumn{2}{c}{Object Detection} & \multicolumn{2}{c}{Instance Segmentation} & \multirow[c]{2}{*}{\shortstack{Trainable\\Params $\downarrow$ (M)}} \\
		& & BD-Rate$\downarrow$ & BD-mAP$\uparrow$ & BD-Rate$\downarrow$ & BD-mAP$\uparrow$ & \\
		\midrule
		Stage II & $ +S $ & -54.063\% & 3.568 & -56.038\% & 3.308 & 0.18 (2.40\%) \\
		& $ +F $ & -52.667\% & 3.335 & -53.160\% & 3.079 & 0.15 (2.00\%) \\
		\midrule
		Stage III & SFA w/o Fusion & -55.635\% & 3.563 & -53.591\% & 3.196 & 0.26 (3.46\%) \\
		& \textbf{SFA} & \textbf{-58.077\%} & \textbf{3.842} & \textbf{-56.042\%} & \textbf{3.421} & 0.40 (5.33\%) \\
		\bottomrule
	\end{tabular}
\end{table*}

\begin{table*}[!htbp]
	\caption{Ablation study of SFA and SCA synergy.}
	\label{ablation_sfa_sca}
	\centering
	\small
	\begin{tabular}{ccccccccccc}
		\toprule
		\multirow[c]{2}{*}{Method} & \multirow[c]{2}{*}{SFA} & \multicolumn{2}{c}{SCA} & \multicolumn{2}{c}{Object Detection} & \multicolumn{2}{c}{Instance Segmentation} & \multirow[c]{2}{*}{\shortstack{Trainable\\Params$\downarrow$ (M)}} \\
		\cline{3-4}
		& &  $h_a$  &  $h_s$  & BD-Rate$\downarrow$ & BD-mAP$\uparrow$ & BD-Rate$\downarrow$ & BD-mAP$\uparrow$ & \\
		\midrule
		(a) & $\checkmark$ &  &  & -58.077\% & 3.842 & -56.042\% & 3.421 & 0.40 (5.33\%) \\
		(b) & $\checkmark$ & $\checkmark$ &  & -59.344\% & 3.891 & -58.385\% & 3.439 & 0.41 (5.50\%) \\
		(c) & $\checkmark$ &  & $\checkmark$ & -58.533\% & 3.931 & -61.271\% & 3.443 & 0.41 (5.50\%)\\
		(d) & $\checkmark$ & $\checkmark$ & $\checkmark$ &  \textbf{-60.824\%} & \textbf{4.014} & \textbf{-61.784\%} & \textbf{3.480} & 0.42 (5.59\%) \\
%		(e) & Adapt-ICMH \cite{Li2024} & $\checkmark$ & $\checkmark$ & -57.894\% & 3.894 & -57.851\% & 3.356 & 0.30 (3.99\%) \\
		(e) & Adapt-ICMH-128 \cite{Li2024} & $\checkmark$ & $\checkmark$ & -58.399\% & 3.709 & -60.936\% & 3.413 & 0.64 (8.52\%) \\
		\bottomrule
	\end{tabular}
\end{table*}

\subsection{Ablation Studies}

\subsubsection{Effect on the architecture of SFA}

\setcounter{table}{6}
\begin{table}[]
	\caption{Effect of varied dimension $C'$ and reduction ratio $r$.}
	\label{middle_r_ablation}
	\centering
	\small
	\begin{tabular}{ccccc}
		\toprule
		\multirow{2}{*}{$C'$} & \multirow{2}{*}{$ r $} 
		& \multicolumn{2}{c}{Object Detection} 
		& \multirow[c]{2}{*}{\shortstack{Trainable Params\\$\downarrow$(M)}} \\
		& & BD-Rate$\downarrow$ & BD-mAP$\uparrow$ & \\
		\midrule
		32 & 16 & -57.622\% & 3.875 & 0.29 (3.86\%) \\
		64 & 8  & -60.824\% & 4.014 & 0.42 (5.59\%) \\
		96 & 4  & -61.329\% & 4.063 & 0.57 (7.59\%) \\
		\bottomrule
	\end{tabular}
\end{table}

\begin{table}[h]
	\caption{Comparison of computational complexity.}
	\centering
	\small
	\setlength{\tabcolsep}{3.5pt} 
	\begin{tabular}{ccccccc}
		\toprule
		\multirow{2}{*}{Model} & \multicolumn{2}{c}{KMACs/pixel} & \multicolumn{2}{c}{Latency (ms)} & \multirow[c]{2}{*}{Params} & \multirow[c]{2}{*}{BD-} \\
		\cmidrule(lr){2-3} \cmidrule(lr){4-5} 
		& Enc. & Dec. & Enc. & Dec. &  $\downarrow$ (M) & Acc $\uparrow$ \\
		\midrule
		full fine-tuning & 132.6 & 176.4 & 94.9 & 94.8 & 7.51 & 17.6 \\
		Ours & 153.6 & 197.3 & 109.5 & 101.0 & 0.42 & 17.4 \\
		\bottomrule
	\end{tabular}
	\label{complexity_comparison_compact}
\end{table}

We ablate the key architectural components of the SFA to quantify their respective contributions, with results presented in \cref{sfa_ablation}. An analysis of the individual branches indicates that both the spatial-only ($ +S $) and frequency-only ($ +F $) branches provide significant gains, confirming their effectiveness. Critically, we find that a naive additive fusion (SFA w/o Fusion) is suboptimal; it even underperforms the spatial-only branch ($ +S $) in the segmentation task. In contrast, the full SFA, which employs a soft fusion mechanism, consistently outperforms all ablated variants. This result empirically demonstrates that the deep, non-linear fusion is not just a minor improvement but is essential for dynamically integrating the spatial and frequency domains to achieve optimal performance.

\subsubsection{Synergy between SFA and SCA}
\cref{ablation_sfa_sca} highlights the synergy between SFA and SCA. While adding SCA to either the entropy encoder \(h_a\) (b) or decoder \(h_s\) (c) on top of the SFA baseline improves performance, the best gains emerge when both are concurrently applied (d). Additionally, replacing SFA with the larger-parameter adapter from Adapt-ICMH \cite{Li2024} severely degrades  performance, demonstrating both our superior structural design and more robust synergy with SCA. The comparison from SFA-only (a) to the full S$^2$-CoT framework (d) validates the necessity of structure-semantics co-tuning. Further analysis confirming the effectiveness of SCA, including its impact on inter-channel and spatial correlation and the elimination of the $\Delta R$ penalty, is in \cref{subsec:supCorrelationRedundancyAnalysis}.

\subsubsection{Placement sensitivity of SFA and SCA}
\begin{figure*}[!htbp]
	\centering
	\begin{tabular}{cccccc}
		\includegraphics[width=0.3\columnwidth]{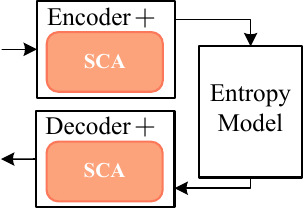} &
		\includegraphics[width=0.3\columnwidth]{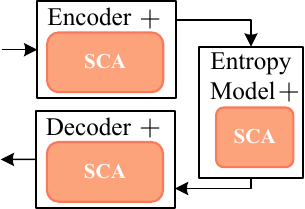} &
		\includegraphics[width=0.3\columnwidth]{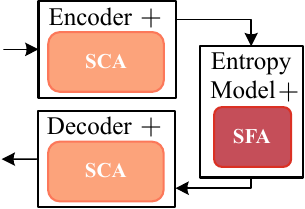} &
		\includegraphics[width=0.3\columnwidth]{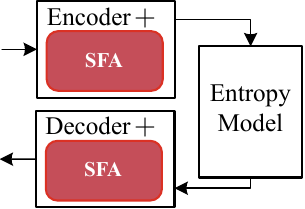} &
		\includegraphics[width=0.3\columnwidth]{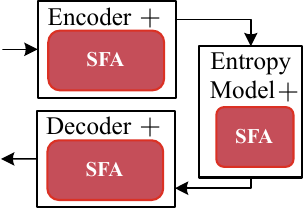} &
		\includegraphics[width=0.3\columnwidth]{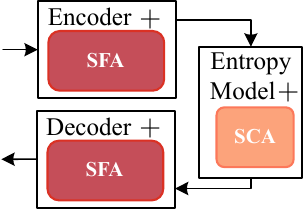} \\
		(a) & (b) & (c) & (d) & (e) & (f)
	\end{tabular}
	\caption{Visual illustration of the six module placement strategies (a)–(f). }
	\label{pos_vs}
\end{figure*}

To validate the role-aware placement strategy, we investigate the sensitivity of SFA and SCA to their architectural placement. The results are reported in \cref{codec_entropy_ablation}, and the various placement strategies are shown in \cref{pos_vs}.

\setcounter{table}{5}
\begin{table*}[!htbp]
	\caption{Ablation study on the placement sensitivity of the SFA and SCA.}
	\label{codec_entropy_ablation}
	\centering
	\small
	\begin{tabular}{cccccccc}
		\toprule
		\multirow[c]{2}{*}{Method} & \multirow[c]{2}{*}{\shortstack{Adapters in\\ Encoder-Decoder}} & \multirow[c]{2}{*}{\shortstack{Adapters in\\Entropy Model}}
		& \multicolumn{2}{c}{Object Detection} 
		& \multicolumn{2}{c}{Instance Segmentation} 
		& \multirow[c]{2}{*}{\shortstack{Trainable\\Params$\downarrow$ (M)}} \\
		& & & BD-Rate$\downarrow$ & BD-mAP$\uparrow$ & BD-Rate$\downarrow$ & BD-mAP$\uparrow$ & \\
		\midrule
		(a) & SCA & --   & -51.179\% & 3.022 & -48.209\% & 2.863 & 0.05 (0.67\%) \\
		(b) & SCA & SCA  & -50.975\% & 3.255 & -51.417\% & 2.983 & 0.07 (0.93\%) \\
		(c) & SCA & SFA  & -49.972\% & 3.160 & -50.445\% & 2.953 & 0.19 (2.53\%) \\
		(d) & SFA & --   & -58.077\% & 3.842 & -56.042\% & 3.421 & 0.40 (5.33\%) \\
		(e) & SFA & SFA  & -55.415\% & 3.977 & -60.365\% & 3.464 & 0.54 (7.19\%) \\
		(f) & \textbf{SFA} & \textbf{SCA}  & \textbf{-60.824\%} & \textbf{4.014} & \textbf{-61.784\%} & \textbf{3.480} & 0.42 (5.59\%) \\
		\bottomrule
	\end{tabular}
\end{table*}
\setcounter{table}{8}

%An analysis of the main encoder-decoder pathway reveals that SFA is the superior adapter for this component. As shown by the comparison in (a) and (d), integrating SFA into the backbone yields a markedly better rate-performance trade-off than placing SCA in the same position. This result confirms that SFA's design is inherently suited for modulating the spatial and frequency domains of the feature representation.
Analysis of the main encoder-decoder pathway reveals that SFA is the superior adapter for this component, as the comparison between configurations (a) and (d) yields a markedly better rate-performance trade-off for SFA.

%Analysis of the main encoder-decoder pathway reveals that SFA is the superior structural adaptation module for this component, as the comparison between configurations (a) and (d) yields a markedly better rate-performance trade-off for SFA, validating its spatial-frequency specialization.

Conversely, for the entropy model, SCA demonstrates clear superiority. The comparison of configurations (e) and (f) shows a substantial performance gap in favor of SCA, underscoring its specialized design for refining statistical context. Configurations (b) and (c) exhibit the same trend.

These results converge to a clear conclusion: optimal performance is achieved only when each adapter is placed in its specialized role, with SFA in the codec and SCA in the entropy model. This empirically validates our core ``what and where" principle of specialized, synergistic adaptation.

\subsubsection{Analysis of adapter hyperparameters}
We analyze the capacity of adapters by varying the SFA's middle dimension $C'$  and the SCA's reduction ratio $r$, with results reported in \cref{middle_r_ablation}. While increasing the middle dimension $C'$ from 32 to 64 yields substantial performance gains, a further increase to 96 results in marginal improvements that do not justify the added parameter complexity. Hence, we select the configuration with $C'$=64 and  $r$=8, as it strikes an optimal balance between task performance and efficiency. \cref{subsec:supDetailedHyperparametersAnalysis} provides a more detailed hyperparameter analysis, with controlled ablations on $C'$ and $r$.

\subsubsection{Computational complexity and efficiency}
The S$^2$-CoT achieves high efficiency with a 94.4\% reduction in trainable parameters. This minimal parameter set introduces negligible overhead, adding only +21 KMACs/pixel and +10 ms latency over the baseline (see \cref{complexity_comparison_compact}). Full complexity and efficiency are provided in \cref{sec:supComputationalComplexityAndEfficiency}.

\subsection{Qualitative Results}
\begin{figure}[!h]
	\centering
	\includegraphics[width=\linewidth]{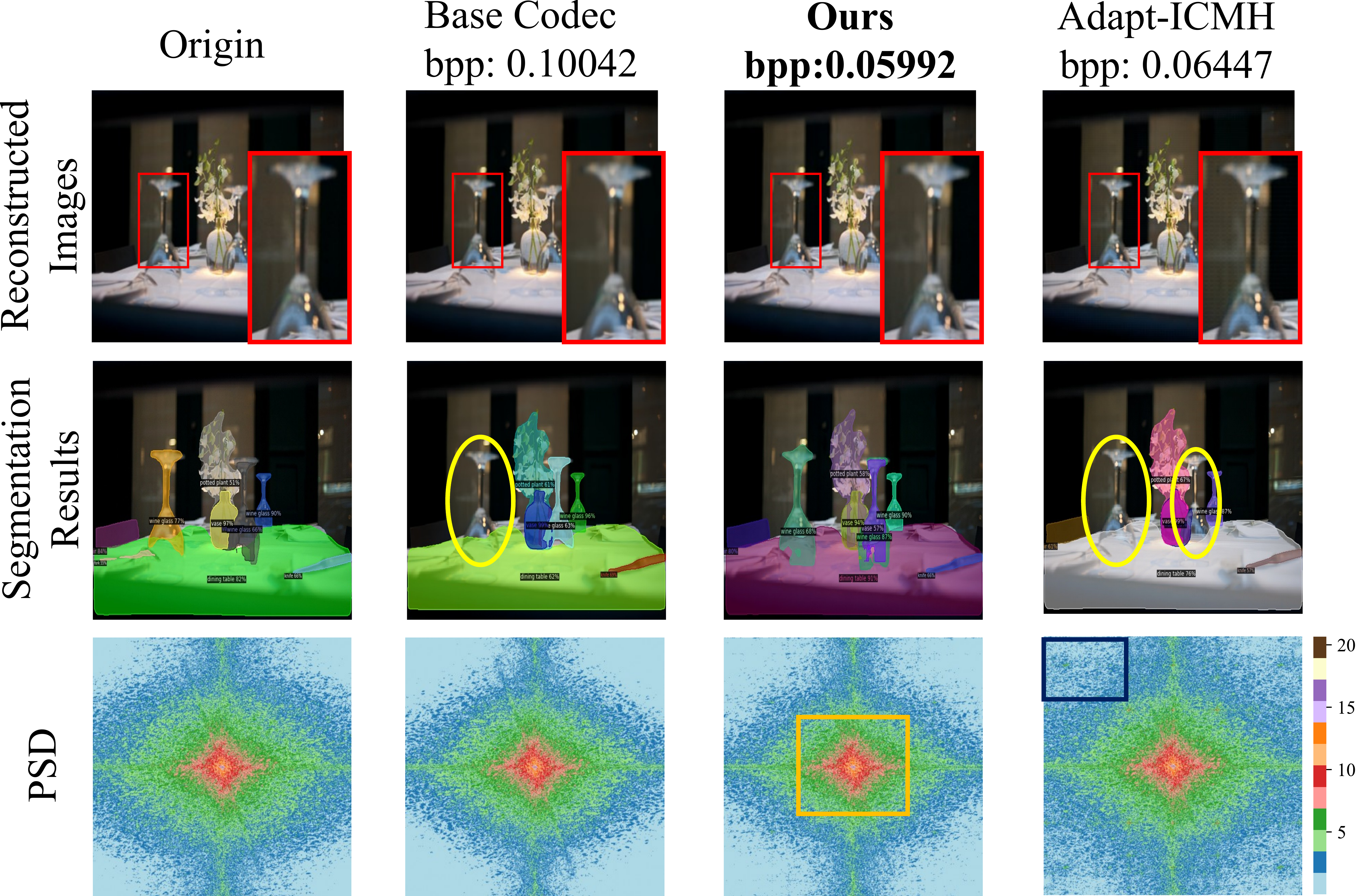} % Reduce the figure size so that it is slightly narrower than the column. Don't use precise values for figure width.This setup will avoid overfull boxes.
	\caption{Qualitative comparison on the COCO2017 dataset.}
	\label{qualitative_results}
\end{figure}
\cref{qualitative_results} shows a qualitative comparison, where our S$^2$-CoT achieves superior visual fidelity and instance segmentation at a lower bitrate. The reconstructed images exhibit enhanced structural coherence and naturalness, maintaining texture fidelity in zoomed-in regions. This improvement is attributed to SFA’s soft fusion mechanism, which selectively preserves critical spatial and high-frequency features, enabling more accurate downstream predictions. As highlighted by the yellow ovals, competing methods miss key foreground instances that our method robustly segments.

The log Power Spectral Density (PSD) analysis further confirms the superiority of our S$^2$-CoT. Its PSD, marked by the orange box, is more concentrated and pure, reflecting effective suppression of low-frequency redundancy while preserving structure-relevant frequencies. In contrast, Adapt-ICMH injects spurious high-frequencies (blue box), disrupting texture integrity and semantic consistency.
\subsection{Application to Classification and Larger Codec}
\subsubsection{Classification Task}
We also experimented with S$^2$-CoT on the ImageNet classification task, with results presented in \cref{main_comparison}(b). Compared to other PEFT methods, our method achieves consistent gains in Top-1 accuracy while maintaining favorable rate-distortion trade-offs. This finding demonstrates the potential of the synergistic adaptation strategy for effective application across a broader range of vision tasks.
\subsubsection{Larger Transformer-based Image Codec}
%We further validate S²-CoT's effectiveness by transferring it to $ \textit{DCAE} $ (CVPR\textquotesingle25) \cite{Lu2025}, a recent Transformer-based codec with over 100M parameters that incorporates a dictionary-based cross attention entropy model. As shown in Figure \cref{main_comparison} (c), S²-CoT consistently maintains robust rate-distortion efficiency and downstream detection performance, even under expanded model capacity. These results underscore S²-CoT's robustness across varied codec scales and highlight its significant potential for seamless integration and performance gains across diverse and larger modern base codecs. More quantitative results are in the appendix.
To further validate the scalability of our framework, we deployed S$^2$-CoT on $ \textit{DCAE} $ (CVPR\textquotesingle25) \cite{Lu2025}, a recent, large-scale (100M+ parameters) Transformer-based codec. It features a hybrid entropy model that enhances a standard hyperprior and channel-wise autoregressive architecture with a dictionary-based cross-attention mechanism. As shown in \cref{main_comparison}(c), S$^2$-CoT again establishes a superior rate-accuracy trade-off, demonstrating robust performance even when paired with a considerably more complex codec. This result confirms that our synergistic adaptation strategy is not only robust but also highly generalizable, highlighting its potential for seamless integration with diverse, SOTA base codecs. More quantitative results are shown in \cref{sec:supMoreQualitativeResultes}.

%% file: sec/5_conclusion.tex
\section{Conclusion}
\label{sec:conclusion}

We rethink parameter-efficient adaptation for LIC and reveal an important but overlooked insight: effective tuning requires coordinated adaptation across both feature and statistical spaces. Our work answers the core question of what and where to adapt by proposing S$^2$-CoT, a dual-adapter framework composed of two specialized, synergistic components: an advanced SFA integrated into the encoder-decoder to enhance structural fidelity, and an SCA, the first PEFT module designed to refine the semantic context within the entropy model by modeling channel interactions. Extensive experimental results on four diverse base codecs confirm that structure–semantics co-tuning is critical, enabling a model-agnostic dual-adapter paradigm for LIC, achieving performance that narrows the gap to full fine-tuning with a small fraction of trainable parameters. Future work will extend our proposed theory and methodology to video and multi-task compression scenarios.

%% file: sec/X_suppl.tex
\clearpage
\setcounter{page}{1}
\maketitlesupplementary

\renewcommand{\thesection}{\Alph{section}} 
\setcounter{section}{0}

%\item The supplementary can back-reference sections of the main paper, for example, we can refer to \cref{sec:intro};
%\item The main paper can forward reference sub-sections within the supplementary explicitly (e.g. referring to a particular experiment); 

%This supplementary material contains additional some discussions in the following sections:
\noindent This supplement provides additional discussion on:

\begin{itemize}
	\item \cref{sec:supPeft}: PEFT in ICMH.
	\item \cref{sec:supWork}: More Related Work.
	\begin{itemize} % [label=\textbullet]
		\item \cref{subsec:supTraditionalImageCompression}: Traditional Image Compression.
		\item \cref{subsec:supLearnedImageCompression}: Learned Image Compression.
	\end{itemize}
	\item \cref{sec:supTheoreticalAnalysis}: Theoretical Analysis and Empirical Analysis of the Structure-Semantics Synergy.
	\begin{itemize} % [label=\textbullet]
		\item \cref{subsec:supDerivationCodingRedundancyPenalty}: Derivation of Coding Redundancy Penalty.
		\item \cref{subsec:supConsequenceIsolatedAdaptation}: Consequence of Isolated Adaptation.
		\item \cref{subsec:supSynergySCA}: Synergy via Semantic Context Adapter.
		\item \cref{subsec:supCorrelationRedundancyAnalysis}: Correlation \& Redundancy Analysis.
	\end{itemize}
	\item \cref{sec:supTaskPerceptualDistortionLoss}: Task-specific Perceptual Distortion Loss.
	\begin{itemize}
		\item \cref{subsec:supLossClassificationTask}: Loss for Classification Task.
		\item \cref{subsec:supLossDetSegTasks}: Loss for Detection and Segmentation Tasks.
	\end{itemize}
	\item \cref{sec:supExperimentalSettingsOfTraining}: Experimental Settings of Training.
	\begin{itemize}
		\item \cref{subsec:supExperimentalEquipment}: Experimental Equipment.
		\item \cref{subsec:supHyperparameterSettings}: Hyperparameter Settings.
	\end{itemize}
	\item \cref{sec:supMoreMainExperimentalResults}: More Main Experimental Results.
	\item \cref{sec:supDetailedAblationStudies}: Detailed Ablation Studies.
	\begin{itemize}
		\item \cref{subsec:supDetailedAblationAdaptICMHSCA}: Detailed Ablation on Adapt-ICMH and SCA.
		\item \cref{subsec:supDetailedAblationStudyonSCAPlacement}: Detailed Ablation Study on SCA Placement.
		\item \cref{subsec:supDetailedHyperparametersAnalysis}: Detailed Hyperparameters Analysis.
	\end{itemize}
	\item \cref{sec:supGeneralizationYOLODetectors}: Generalization to YOLO Detectors.
	\item \cref{sec:supComputationalComplexityAndEfficiency}: Computational Complexity and Efficiency.
	\item \cref{sec:supIntegrationDetailsOfAdapters}: Integration Details of Adapters.
	\item \cref{sec:supMoreResultsOnTheDiverseBaseCodec}: More Results on the Diverse Base Codec.
	\item \cref{sec:supQuantitativeResultsOnClassificationTask}: Quantitative Results on Classification Task.
	\item \cref{sec:supSensitivityNecessityOfCo-Tuning}: Sensitivity \& Necessity of Co-Tuning.
	\item \cref{sec:supFrameworkParadigmAndModularity}: Framework Paradigm and Modularity.
	\item \cref{sec:supPyTorchImplementationOfSFAAndSCA}: PyTorch Implementation of SFA and SCA.
	\item \cref{sec:supFutureWork}: Future Work.
	\item \cref{sec:supMoreQualitativeResultes}: More Qualitative Results.
	\item \cref{sec:supgeneralizationSynergy}: Generalization of Synergy to Advanced Entropy Models Across Diverse Base Codecs.
	\begin{itemize} % [label=\textbullet]
		\item \cref{subsec:supSynergyInAutoregressiveGMMs}: Synergy in Autoregressive GMMs.
		\item \cref{subsec:supSynergyInDictionaryEntropyModels}: Synergy in Dictionary Entropy Models.
		\item \cref{subsec:supContextSynergyInSpatial-Channel}: Context Synergy in Spatial-Channel.
		\item \cref{subsec:supUnifiedPerspective}: Unified Perspective.
	\end{itemize}
\end{itemize}

\section{PEFT in ICMH}
\label{sec:supPeft}
\begin{table}[htbp]
	\centering
	\caption{Performance comparison on different tasks of ICMH.}
	\label{psnr_com}
	\setlength{\tabcolsep}{1.0pt}
	\small
	\begin{tabular}{cccccc}
		\toprule
		\multirow{2}{*}{Tasks} & \multirow{2}{*}{Models} & \multirow{2}{*}{\shortstack{Bit\\Rate}} & \multirow{2}{*}{mAP ($\uparrow$)} & \multirow{2}{*}{PSNR ($\uparrow$)} & \multirow{2}{*}{\shortstack{PSNR w/o\\Adapters ($\uparrow$)}} \\
		&&&&&\\
		\midrule
		\multirow{2}{*}{\shortstack{object\\detection}} 
		& base codec & 0.056 & 27.123 & 30.286 & 30.286 \\
		& S$^2$-CoT & 0.0814 & 36.767 & 27.044 & 30.286 \\
		\midrule
		\multirow{2}{*}{\shortstack{instance\\segmentation}} 
		& base codec & 0.056 & 24.81 & 30.29 & 30.29 \\
		& S$^2$-CoT & 0.103 & 34.64 & 28.18 & 30.29 \\
		\bottomrule
	\end{tabular}
\end{table}
We introduce a paradigm shift in Learned Image Compression (LIC) by incorporating Parameter-Efficient Fine-Tuning (PEFT) to resolve the scalability bottleneck in multi-task deployment. While traditional full fine-tuning yields task-specific gains, it necessitates maintaining separate, heavyweight checkpoints for every downstream application, rendering deployment prohibitively expensive in terms of storage and version management. In contrast, our plug-and-play adapter design offers a superior performance-flexibility trade-off. This architecture disentangles optimization goals: the frozen base codec preserves optimal perceptual quality for human viewing, while the insertion of lightweight adapters maximizes machine-task accuracy. Consequently, this enables rapid, low-cost adaptation to diverse analytics scenarios, where maintaining a library of tiny adapters is orders of magnitude more efficient than storing $N$ independent, fully fine-tuned models.

To clarify the benefits of our task-oriented modules and the associated trade-offs, we provide a detailed quantitative analysis in \cref{psnr_com}. Our S$^2$-CoT framework employs Parameter-Efficient Fine-Tuning adapters to decouple machine vision optimization from human perceptual requirements. As shown in \cref{psnr_com}, activating the adapters significantly boosts downstream performance, exemplified by substantial gains in object detection metrics, with only a marginal increase in bitrate. While optimizing specifically for machine accuracy involves a trade-off in pixel-wise fidelity as reflected in the decrease of PSNR, this aligns with the consensus that semantic feature preservation often diverges from pixel-perfect reconstruction. Crucially, the primary benefit of introducing these extra modules lies in scalability and flexibility. As evidenced by the ``PSNR w/o Adapters" column, our framework maintains the capability to revert to the optimal human-perceptual quality of the base codec simply by deactivating the adapters. This design allows a single system to support diverse transmission demands: it provides high-efficiency feature coding for machine analytics when adapters are active and high-fidelity reconstruction for human viewing when bypassed, achieving the best of both worlds without altering the codec.

\section{More Related Work}
\label{sec:supWork}

This section supplements the main paper's related work by providing a broader historical and technical context. We first review traditional codecs to establish a historical baseline, followed by a summary of the general evolution of LIC. This provides the foundational context for the more specialized methods discussed in the main paper.

\subsection{Traditional Image Compression}
\label{subsec:supTraditionalImageCompression}

Traditional image compression methods, such as JPEG \cite{Wallace1991} and JPEG2000  \cite{Christopoulos2000, BartrinaRapesta2009}, rely on discrete cosine transform and wavelet decomposition to exploit perceptual redundancy. The objective of optimization is to reduce image distortion. These methods have been successful in removing spatial redundancy from images and are able to satisfy the image quality requirements of human vision to a certain extent. However, given that these systems were not originally designed for machine vision tasks, applying them to such contexts results in a serious degradation of downstream per-\\formance due to the failure to preserve semantic integrity.

\subsection{Learned Image Compression}
\label{subsec:supLearnedImageCompression}

Learned image compression \cite{Balle2017, Minnen2018, Balle2018, Cheng2020, Zou2022, Han2024, Ji2025, Li2024a, Liu2023b, Zeng2025} has revolutionized the field by employing advanced end-to-end optimized neural networks to surpass the rate-distortion performance of traditional codecs \cite{Wallace1991, Christopoulos2000, BartrinaRapesta2009}. Foundational works by Ballé \etal \cite{Balle2017, Balle2018} established this paradigm, demonstrating the power of learning nonlinear transforms from data.

Subsequent research has advanced LIC by exploring increasingly powerful architectures for the nonlinear transform backbone. Architectures have evolved from early CNN-based designs \cite{Cheng2020} to more powerful Transformer-based models \cite{Liu2023, Zou2022} that better capture both local and global dependencies in the image. Despite these significant architectural advancements, the majority of these foundational LIC methods are optimized for human-centric metrics (\eg, PSNR, MS-SSIM). Consequently, they exhibit limited adaptability for diverse machine vision tasks, which prioritize the preservation of task-critical semantic, rather than pixel-level, fidelity.

\section{Theoretical Analysis and Empirical Analysis of the Structure-Semantics Synergy}
\label{sec:supTheoreticalAnalysis}

This section complements Sec. 3.1 of the main paper by providing a comprehensive and rigorous theoretical derivation of the coding redundancy penalty ($\Delta R$) and presenting a detailed analysis of the synergy between the Structural Fidelity Adapter (SFA) and the Semantic Context Adapter (SCA). We demonstrate how the proposed S$^2$-CoT framework effectively aligns the structural representation with statistical modeling to minimize redundancy.

\subsection{Derivation of Coding Redundancy Penalty}
\label{subsec:supDerivationCodingRedundancyPenalty}

In a standard LIC framework, the non-linear encoder-decoder (also called transform) $(g_{a}, g_{s})$ and the entropy model $(h_{a}, h_{s})$ are jointly optimized. The encoder $g_a$ maps an input $x$ to a compact latent representation $y$, while the entropy model estimates the probability distribution $p_{\hat{y}}$ of its quantized version $\hat{y}$ to minimize the bitrate.

In hyperprior-based models \cite{Balle2018}, the latent $y$ is typically modeled as a Gaussian distribution conditioned on the hyper-latent $\hat{z}$:
\begin{equation}
	p_{\hat{y}|\hat{z}}(\hat{y}|\hat{z}) \sim \mathcal{N}(\mu, \sigma^{2})
\end{equation}
where the Gaussian parameters $(\mu, \sigma)$ are predicted by the hyperprior decoder $h_s$:
\begin{equation}
	(\mu, \sigma) = h_{s}(\hat{z}; \theta_{h_{s}}), \quad \text{with} \quad \hat{z} = Q(h_{a}(y; \theta_{h_{a}}))
\end{equation}
The theoretical bitrate $R(\hat{y})$ is given by the cross-entropy between the marginal distribution of the latents and the estimated probability model:
\begin{equation}
	R(\hat{y}) = \mathbb{E}_{\hat{y} \sim p_{\text{data}}} [-\log_{2}(p_{\hat{y}|\hat{z}}(\hat{y}|\hat{z}))]
\end{equation}

\subsection{Consequence of Isolated Adaptation}
\label{subsec:supConsequenceIsolatedAdaptation}

When we perform Parameter-Efficient Fine-Tuning (PEFT) using only the structural adapter in the encoder-decoder $(g_a, g_s)$ while keeping the entropy model frozen \cite{Li2024}, the latent representation shifts from $y$ to a task-adapted feature space $y'$. This shift alters the statistical properties of the distribution, denoted as $p_{\text{data}}'$.

However, since the entropy model parameters $(\theta_{h_a}, \theta_{h_s})$ remain frozen, the predicted Gaussian parameters $(\mu', \sigma')$ are derived from the pre-trained (human-oriented) priors:
\begin{equation}
	(\mu', \sigma') = h_{s}(\hat{z}'; \theta_{h_{s}}), \quad \text{where} \quad \hat{z}' = Q(h_{a}(y'; \theta_{h_{a}}))
\end{equation}
These parameters $(\mu', \sigma')$ are suboptimal for the new distribution $y'$, creating a deviation from the ideal parameters $(\mu_{\text{ideal}}, \sigma_{\text{ideal}})$ that would model the task-adapted latents. This statistical misalignment leads to a non-negligible coding redundancy penalty $\Delta R$, defined as the Kullback-Leibler (KL) divergence between the ideal distribution and the frozen model's estimation.

Substituting this into the bitrate equation (Eq. (2) in the main paper), we derive:
\begin{align}
	R(\hat{y}') &= \mathbb{E} [-\log_{2}(p_{\text{frozen}}(\hat{y}'|\hat{z}'))] \nonumber \\
	&= \underbrace{\mathbb{E} [-\log_{2}(p_{\text{ideal}}(\hat{y}'|\hat{z}'))]}_{\text{Ideal Bitrate}} + \underbrace{D_{\text{KL}}(p_{\text{ideal}} || p_{\text{frozen}})}_{\text{Penalty } \Delta R}
\end{align}
Here, $\Delta R > 0$ represents the extra bits consumed due to the failure of the frozen entropy model to capture the structural changes introduced by the transform-only adapter.
\begin{figure}[!htbp]
	\centering
	\includegraphics[width=\linewidth]{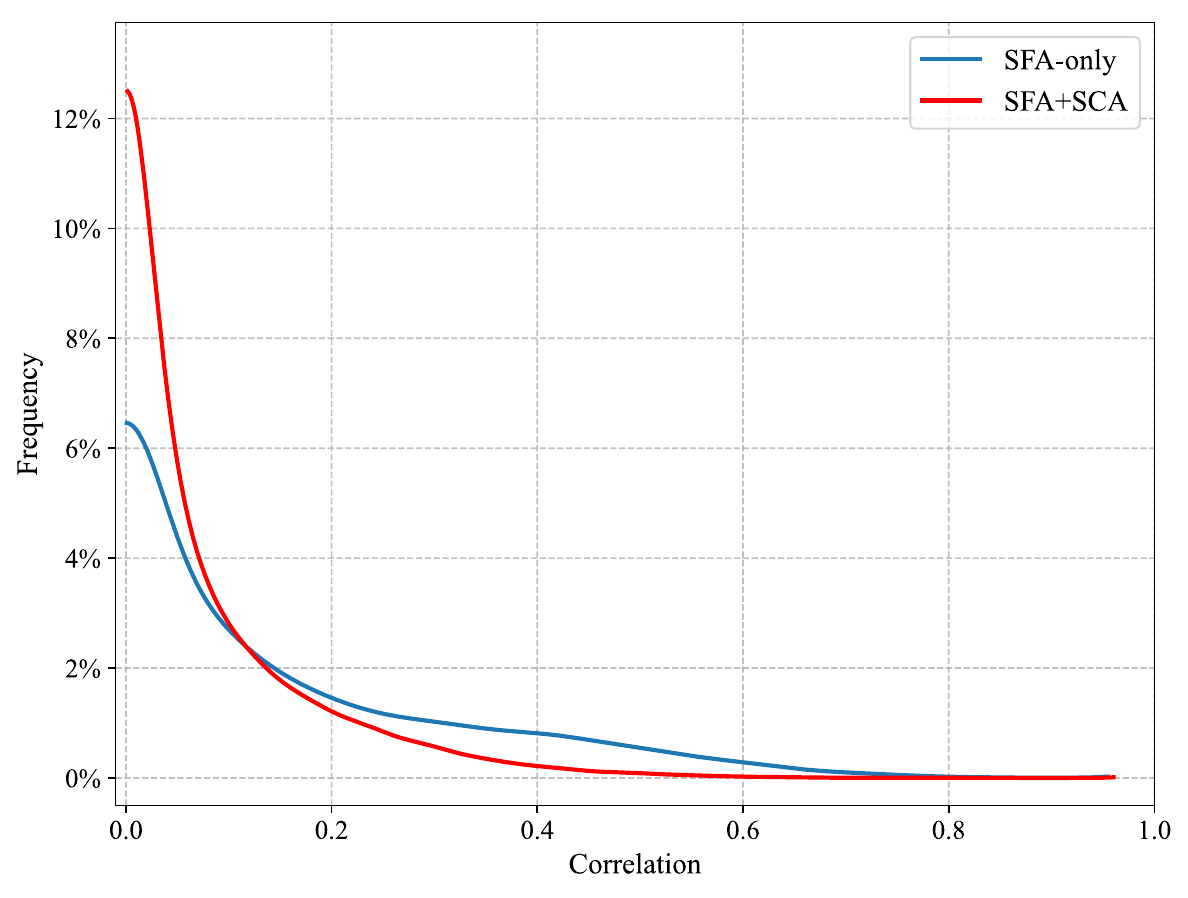}
	\caption{ Pairwise channel similarities of the latent representation $y$ for different fine-tuning strategies. Incorporating SCA markedly reduces the inter-channel correlation.}
	\label{cor_map1}
\end{figure}

\subsection{Synergy via Semantic Context Adapter}
\label{subsec:supSynergySCA}

The proposed context adapter is explicitly designed to eliminate this penalty. By inserting semantic context adapter into the entropy model, we introduce a learnable correction term that adapts the frozen priors to the new latent statistics:
\begin{equation}
	(\mu_{\text{tuned}}, \sigma_{\text{tuned}}) = \text{SCA}(h_{s}(\hat{z}'; \theta_{h_{s}}))
\end{equation}
Through the joint optimization of S$^2$-CoT, the SCA effectively minimizes the KL divergence term, driving $\Delta R \to 0$. This ensures that the structural gains provided by SFA are preserved without incurring a bitrate penalty, realizing the theoretical basis of our structure-semantics synergy.

\subsection{Correlation \& Redundancy Analysis}
\label{subsec:supCorrelationRedundancyAnalysis}

To empirically validate the theoretical derivation above, we analyze the statistical properties of the latent representations produced by different adaptation strategies. We compare the baseline using SFA-only (isolated structural tuning) against our proposed SFA+SCA (S$^2$-CoT) framework.
\subsubsection{Inter-channel correlation}
We first calculate the pairwise channel similarity of the latent $y$ to estimate inter-channel redundancy. High correlation implies that multiple channels encode similar information, wasting bitrate.

As shown in \cref{cor_map1}, the SFA-only baseline exhibits a heavy tail and a low initial value in the similarity distribution, indicating significant unresolved redundancy. In contrast, the SFA+SCA model shows a steeper decay, suggesting that the SCA effectively decorrelates the features. This confirms that the SCA refines the channel context to learn a more compact, orthogonal representation for machine tasks.

\begin{figure*}[!htbp]
	\centering
	\includegraphics[width=1\linewidth]{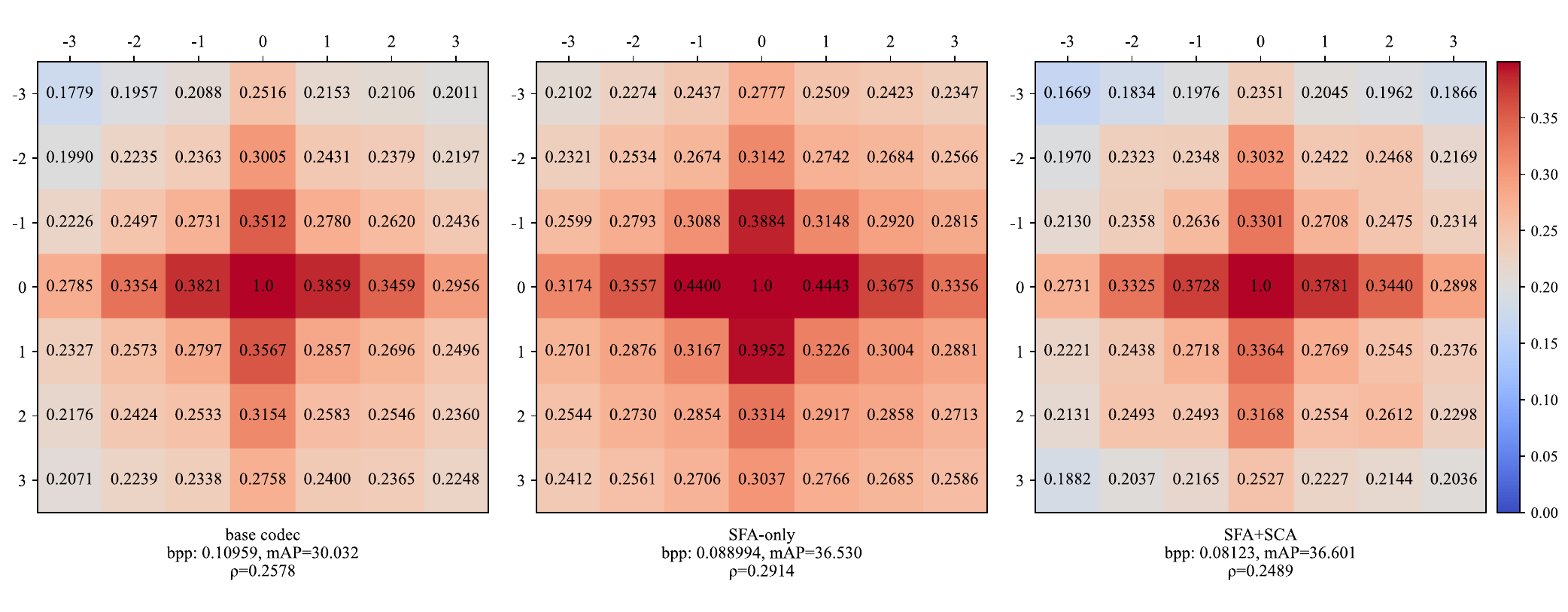}
	\caption{Spatial correlation of $ (y-\mu)/\sigma $ for models trained with $ \lambda $=0.5. SFA+SCA (right) reduces average spatial correlation compared to SFA-only (middle) and the base codec (left),  which benefits subsequent machine vision tasks.}
	\label{cor_spatial_map}
\end{figure*}
\begin{figure*}[!htbp]
	\centering
	\includegraphics[width=1\linewidth]{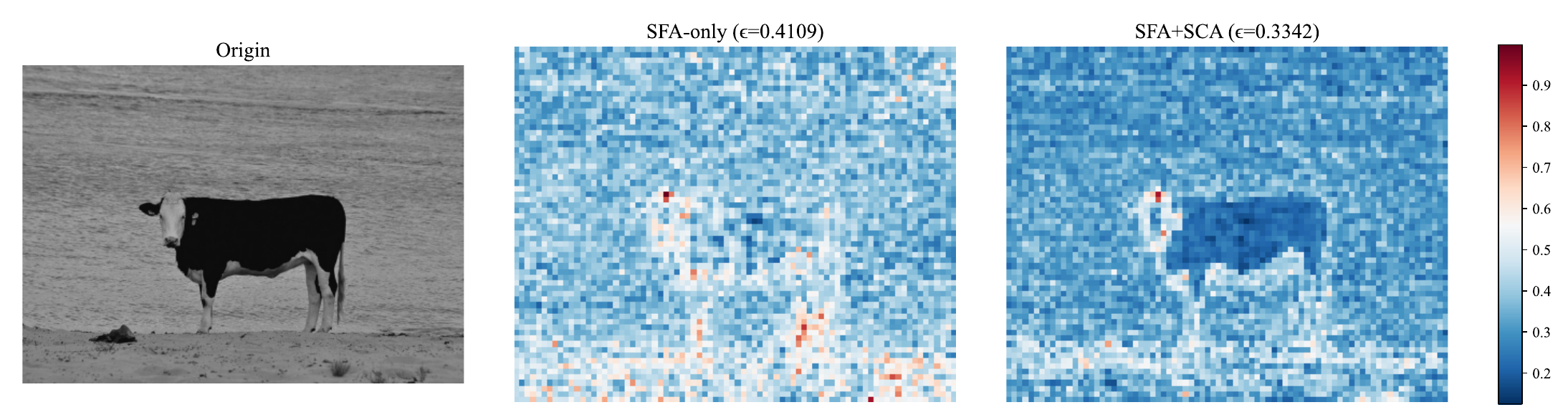}
	\caption{Scaled deviation map of two strategies: SFA-only and SFA+SCA.}
	\label{dev_map1}
\end{figure*}

\subsubsection{Spatial correlation}
We extend the analysis to the spatial dimension by measuring the correlation between adjacent spatial positions in the normalized latents $(y-\mu)/\sigma$. Ideally, an optimized entropy model should whiten the latents, resulting in near-zero spatial correlation.

As visualized in \cref{cor_spatial_map}, the SFA-only strategy results in a high average spatial correlation ($\rho=0.2914$), reflecting the frozen entropy model's inability to capture the new spatial structures. The synergistic SFA+SCA approach substantially reduces this to $\rho=0.2489$. This reduction serves as direct evidence that the S$^2$-CoT framework successfully aligns the statistical model with the structural features.

\subsubsection{Analysis of quantization deviation}
To further evaluate the precision of the probability estimation, we analyze the	``scaling deviation" metric $\epsilon$, which acts as a proxy for quantization loss \cite{Xie2021}. It measures the discrepancy between the latent $y$ and the estimated mean $\mu$:
\begin{equation}
	\epsilon = \frac{|Q(y-\mu) - (y-\mu)|}{\sum |y|}.
\end{equation}
A smaller $\epsilon$ indicates that the entropy model's predicted mean $\mu$ is closer to the actual latent center, minimizing quantization error.

\cref{dev_map1} presents the scaled deviation maps. The SFA-only approach yields high deviation errors (warmer regions), confirming the mismatch where the frozen priors fail to center the task-adapted latents. Conversely, the addition of SCA in our framework significantly suppresses these errors (cooler regions), particularly in the background. This validates that SCA effectively corrects the statistical shift $\Delta \mu$, ensuring precise quantization and efficient coding.

\subsubsection{Visualizing entropy distribution}
\begin{figure*}[!htbp]
	\centering
	\includegraphics[width=0.9\linewidth]{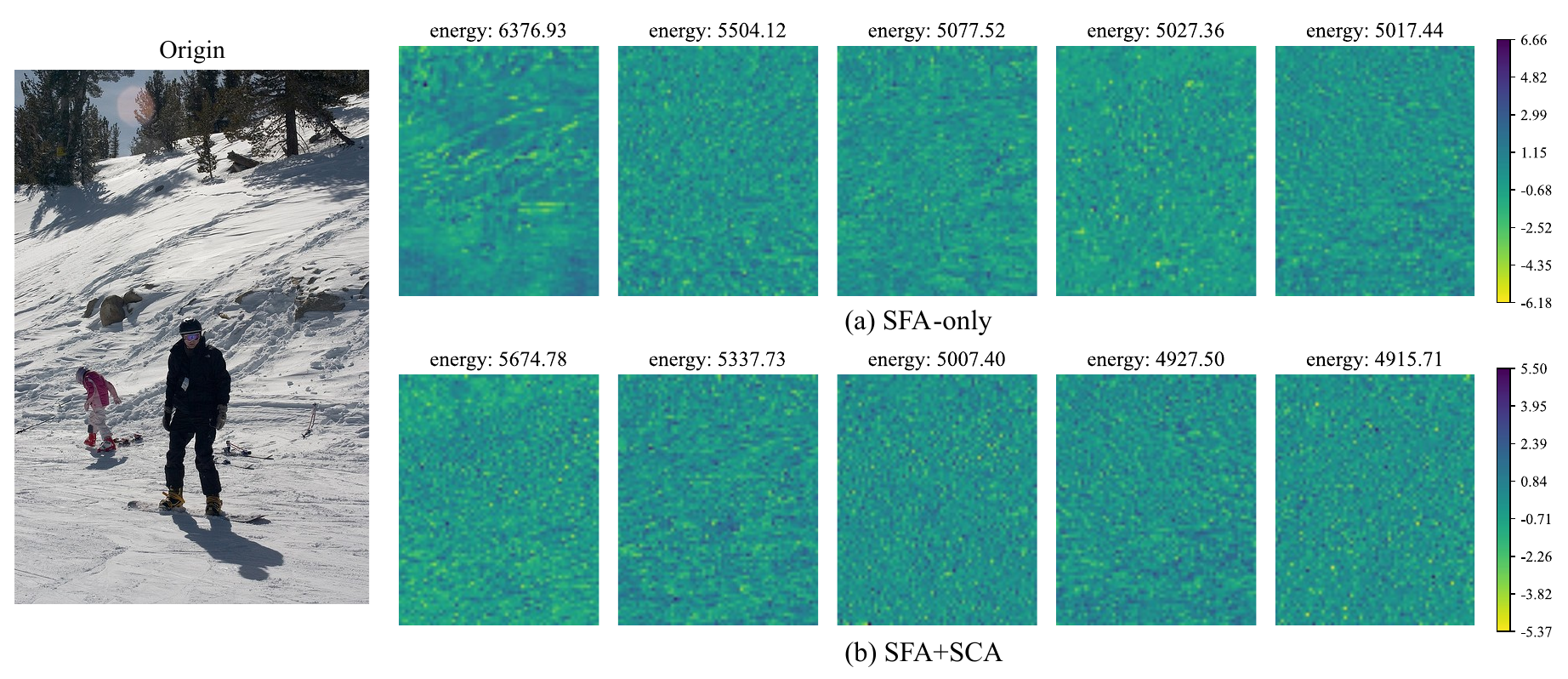}
	\caption{Each row corresponds to a different strategy and shows the energy of five channels with the highest entropy.}
	\label{energy_norm_y}
\end{figure*}

Meanwhile, we provide a qualitative view of the latent representation energy. \cref{energy_norm_y} displays the five channels with the highest entropy. The latents from the SFA+SCA strategy exhibit significantly less perceptible structure (more ``noise-like") compared to SFA-only. In compression theory, a more unstructured, whitened latent representation indicates better decorrelation and higher coding efficiency. This visually confirms that our S$^2$-CoT effectively removes semantic redundancy while preserving the task-critical information in the decoder features.

\begin{figure}[!htbp]
	\centering
	\includegraphics[width=1\linewidth]{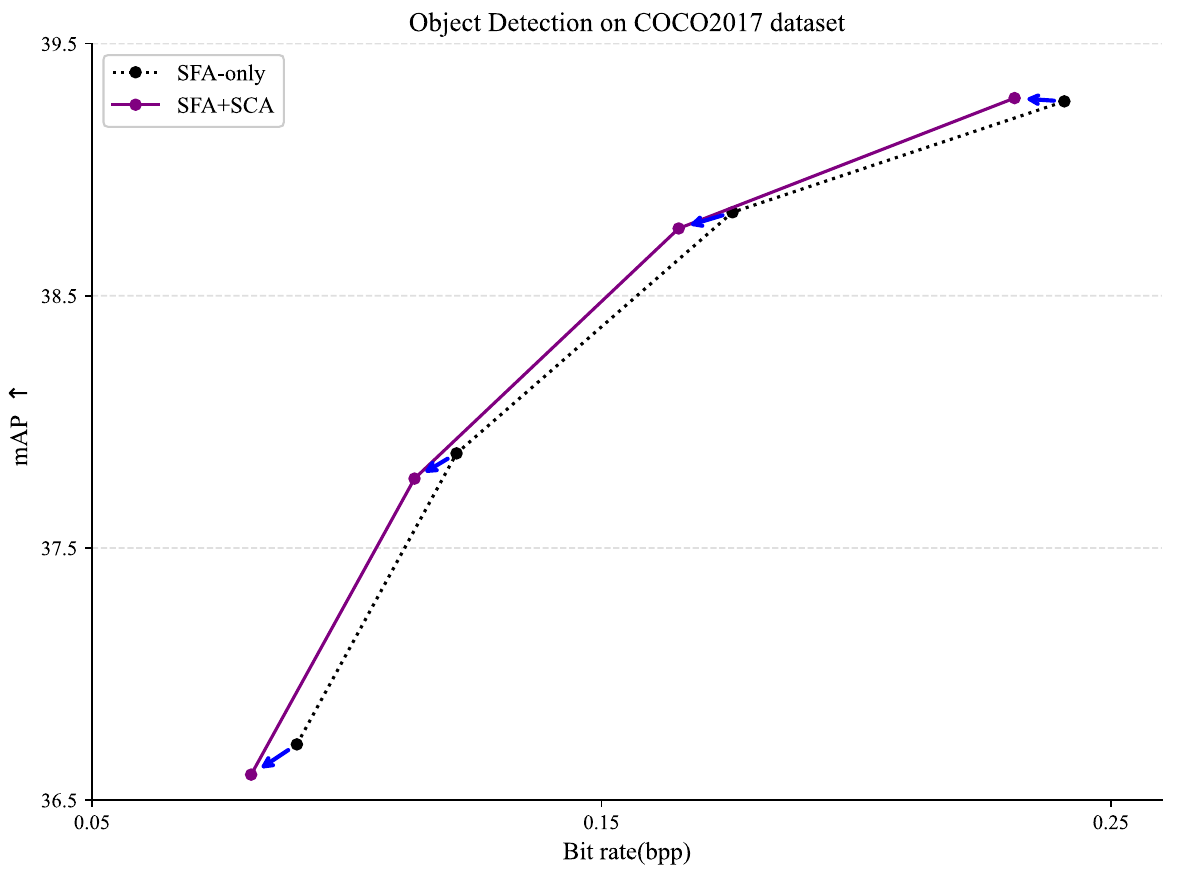}
	\caption{Object detection performance on the COCO2017 dataset using $ \textit{Lu2022-TIC} $ as the base codec. Incorporating SCA reduces the bitrate without compromising detection accuracy.}
	\label{norm_y_bpp}
\end{figure}

\subsubsection{Quantitation of redundancy elimination}
As visualized in the rate-accuracy curve for object detection on COCO2017 (see \cref{norm_y_bpp}), the deployment of our full S$^2$-CoT framework (SFA+SCA) establishes a superior Pareto frontier compared to the isolated SFA-only baseline. The distinct leftward shift, highlighted by the blue arrows, signifies a substantial reduction in bitrate (bpp) at equivalent mean Average Precision (mAP) levels. This empirical evidence validates that the SCA effectively rectifies the statistical misalignment induced by structural fine-tuning, thereby eliminating coding redundancy and yielding a more compact, task-optimal latent representation.

\section{Task-specific Perceptual Distortion Loss}
\label{sec:supTaskPerceptualDistortionLoss}

\begin{equation}
	\mathcal{L} = \mathcal{R}(\hat{\mathbf{y}}) + \lambda \cdot \mathcal{D}(\mathbf{x}, \hat{\mathbf{x}}; \mathcal{G})
	\label{eq_1_loss_function}
\end{equation}
In \cref{eq_1_loss_function} of the main paper, we employ a task-specific perceptual distortion loss, $ \mathcal{D} $, to optimize our adapters for downstream machine vision tasks. A key advantage of this approach is that it enables end-to-end training of the task-specific modules without requiring access to ground-truth task labels (\eg, bounding boxes or segmentation masks).

Our implementation strictly follows the established setup in \cite{Chen2023, Li2024}. Specifically, we utilize a frozen, pre-trained recognition model, $ \mathcal{G} $, as a feature extractor to measure the distance between the original image, $ \mathbf{x} $, and the reconstructed image, $ \hat{\mathbf{x}} $, in the feature space. The choice of $ \mathcal{G} $ depends on the downstream task:
\begin{itemize}
	\item Classification: ResNet50 \cite{He2016}
	\item Object Detection: Faster R-CNN \cite{Ren2015} (with a ResNet50-FPN backbone)
	\item Instance Segmentation: Mask R-CNN \cite{He2017} (with a ResNet50-FPN backbone)
\end{itemize}

\begin{figure}[!htbp]
	\centering
	\includegraphics[width=0.95\columnwidth]{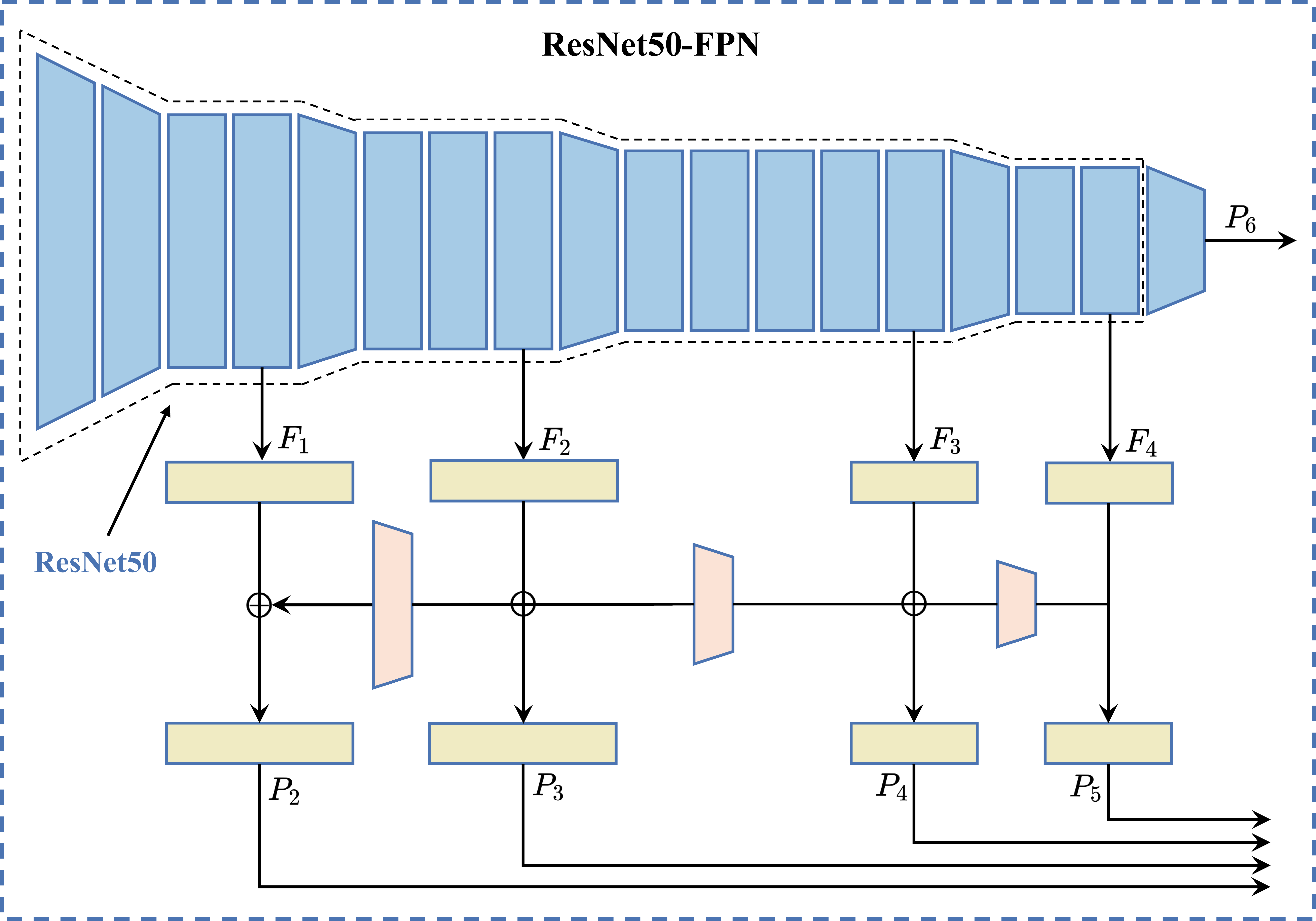} % Reduce the figure size so that it is slightly narrower than the column. Don't use precise values for figure width.This setup will avoid overfull boxes.
	\caption{ResNet50-based FPN architecture, indicating the features selected for perceptual loss evaluation.}
	\label{fig_1_resnet50_fpn}
\end{figure}

\subsection{Loss for Classification Task}
\label{subsec:supLossClassificationTask}

For the classification task, the perceptual distortion is evaluated by computing the Mean Squared Error (MSE) between feature maps extracted from different stages of the ResNet50 backbone, as shown in \cref{fig_1_resnet50_fpn}. We extract the outputs from its four main stages (denoted as $ F_{1} $,$ F_{2} $,$ F_{3} $,$ F_{4} $). The loss is defined as \cref{eq_2_cls_loss_function}:
\begin{equation}
	\mathcal{D}(\mathbf{x}, \hat{\mathbf{x}}; \mathcal{G}) = \frac{1}{4} \sum_{j=1}^{4} \operatorname{MSE}\left(F_{j}(\mathrm{x}), F_{j}(\hat{\mathrm{x}})\right)
	\label{eq_2_cls_loss_function}
\end{equation}
where $ F_{j}(\cdot{}) $ denotes the feature map from the j-th stage of the model $ \mathcal{G} $, ResNet50.

\subsection{Loss for Detection and Segmentation Tasks}
\label{subsec:supLossDetSegTasks}

For dense prediction tasks, object detection, and instance segmentation, the perceptual distortion is evaluated by computing the MSE between feature maps from the Feature Pyramid Network (FPN) of the respective downstream model. We extract the five output levels of the FPN (denoted as $ P_{2} $,$ P_{3} $,$ P_{4} $,$ P_{5} $, $ P_{6}$). The loss is defined as \cref{eq_3_det_seg_loss_function}:
\begin{equation}
	\mathcal{D}(\mathbf{x}, \hat{\mathbf{x}} ; \mathcal{G})=\frac{1}{5} \sum_{j=2}^{6} \operatorname{MSE}\left(P_{j}(\mathbf{x}), P_{j}(\hat{\mathbf{x}})\right)
	\label{eq_3_det_seg_loss_function}
\end{equation}
where $ P_{j}(\cdot{}) $ denotes the feature map from the j-th pyramid level of the FPN. This method effectively measures the fidelity of the multi-scale features that are critical for these downstream tasks.

\section{Experimental Settings of Training}
\label{sec:supExperimentalSettingsOfTraining}

\subsection{Experimental Equipment}
\label{subsec:supExperimentalEquipment}

All experiments were conducted on a single NVIDIA L40 (48GB) or GeForce RTX 4090 (24GB) GPU. Our implementation is based on PyTorch v2.5, and all models were trained using CUDA 12.4.

\subsection{Hyperparameter Settings}
\label{subsec:supHyperparameterSettings}

\cref{tab_1_hyperparam_setting} summarizes the key hyperparameters used for training across all machine vision tasks, namely classification, object detection, and instance segmentation.
\begin{table}[!htbp]
	\centering
	\caption{Training hyperparameter settings for the main trials.}
	\label{tab_1_hyperparam_setting}
	\small
	\setlength{\tabcolsep}{0.6mm}
	\begin{tabular}{cccc} 
		\toprule 
		& Classification & Detection & Segmentation \\ 
		\midrule
		Optimizer & Adam & Adam & Adam \\
		Batch size & 16 & 8 & 8 \\ 
		Epochs & 8 & 40 & 40 \\
		Base learning rate & 1e-4 & 1e-4 & 1e-4 \\
		\midrule
		\multicolumn{4}{c}{$ \textit{Lu2022-TIC} $}\\
		\midrule
		\multirow{2}{*}{\shortstack{Trade-off term\\$\lambda$}} & \multirow{2}{*}{\shortstack{[2.5, 3.5, 5,\\6.7, 13]}} & 
		\multirow{2}{*}{\shortstack{[0.5, 0.9,\\1.8, 3.2]}} &
		\multirow{2}{*}{\shortstack{[0.35, 0.5, 0.9,\\1.8, 3.2]}} \\
		& & & \\
		\midrule
		\multicolumn{4}{c}{$ \textit{\shortstack{Cheng2020-anchor}} $}\\
		\midrule
		\multirow{2}{*}{\shortstack{Trade-off term\\$\lambda$}} & \multirow{2}{*}{-} & 
		\multirow{2}{*}{\shortstack{[0.5, 0.9, 1.8,\\3.2, 5.0]}} &
		\multirow{2}{*}{\shortstack{[0.9, 1.8,\\3.2, 5.0]}} \\
		& & & \\
		
		\midrule
		\multicolumn{4}{c}{$ \textit{\shortstack{DCAE}} $ \& $ \textit{\shortstack{ELIC}} $}\\
		\midrule
		\multirow{2}{*}{\shortstack{Trade-off term\\$\lambda$}} & \multirow{2}{*}{-} & 
		\multirow{2}{*}{\shortstack{[0.5, 0.9,\\1.8, 3.2]}} &
		\multirow{2}{*}{\shortstack{[0.5, 0.9,\\1.8, 3.2]}} \\
		& & & \\
		\bottomrule
	\end{tabular}
\end{table}

\section{More Main Experimental Results}
\label{sec:supMoreMainExperimentalResults}

As referenced in the main paper, we provide a full quantitative comparison for object detection and instance segmentation tasks in \cref{tab_2_cheng2020_results}. The experiments are conducted on the Cheng2020-anchor framework, with the evaluation metrics including BD-Rate savings and BD-mAP gains \cite{Bjoentegaard2001}. Notably, we have included an additional entry, Ours ($C'=4, r=32$), to demonstrate the performance of a more compact version of the S$^2$-CoT. These detailed results provide a comprehensive view of our method's performance and its competitive standing against other methods.

\begin{table*}[!htbp]
	\caption{Extended quantitative results for object detection and instance segmentation tasks based on the $ \textit{Cheng2020-anchor} $ base codec. This table provides the full results that are referenced in the main paper. Best results are in \textbf{bold}, second-best are \underline{underlined}.}
	\label{tab_2_cheng2020_results}
	\centering
	\small
	\begin{tabular}{cccccccc}
		\toprule
		\multirow[c]{2}{*}{Base} & \multirow[c]{2}{*}{Method} & \multirow[c]{2}{*}{Venue} & \multicolumn{2}{c}{Object Detection} & \multicolumn{2}{c}{Instance Segmentation} & \multirow[c]{2}{*}{\shortstack{Trainable\\Params$\downarrow$ (M)}} \\
		%		\cline{4-7}
		& & & BD-Rate$\downarrow$ & BD-mAP$\uparrow$ & BD-Rate$\downarrow$ & BD-mAP$\uparrow$ & \\
		%		\midrule
		%		\multirow[c]{6}{*}{$ \textit{Lu2022-TIC} $}
		%		& full fine-tuning & -- & -73.943\% & 4.511 & -67.977\% & 3.755 & 7.51 (100.00\%) \\ %\cline{2-8}
		%		& Channel Selection & ICPR\textquotesingle22 & 6.849\% & -0.550 & 16.511\% & -0.949 & 0.92 (12.25\%) \\
		%		& TransTIC & ICCV\textquotesingle23 & -46.301\% & 2.768 & -46.521\% & 2.690 & 1.62 (21.57\%) \\
		%		& ICMH-Net & ACM MM\textquotesingle23 & -9.080\% & 0.625 & -10.772\% & 0.654 & 3.98 (53.00\%) \\
		%		& Adapt-ICMH & ECCV\textquotesingle24 & -55.150\% & 3.547 & -52.407\% & 3.208 & \underline{0.29 (3.86\%)} \\
		%		& \underline{Ours ($C'$=32, $r$=32)} & -- & \underline{-60.276\%} & \underline{3.852} & \underline{-60.255\%} & \underline{3.386} & \textbf{0.28 (3.73\%)} \\
		%		& \textbf{Ours} & -- & \textbf{-60.824\%} & \textbf{4.014} & \textbf{-61.784\%} & \textbf{3.480} & 0.42 (5.59\%) \\
		\midrule
		\multirow[c]{7}{*}{ $ \textit{\shortstack{Cheng2020\\-anchor}} $ }
		& full fine-tuning & -- & -59.015\% & 4.699 & -74.052\% & 3.869 & 26.60 (100.00\%) \\ 
		& Channel Selection & ICPR\textquotesingle22 & -11.520\% & 0.723 & -5.156\% & 0.230 & 1.34 (5.04\%) \\
		& ICMH-Net & ACM MM\textquotesingle23 & -8.365\% & 0.503 & -11.980\% & 0.737 & 4.43 (16.65\%) \\
		& Adapt-ICMH & ECCV\textquotesingle24 & -49.245\% & 3.121 & \underline{-60.657\%} & \underline{3.389} & 0.41 (1.54\%) \\
		& SVD-LoRA & CVPR\textquotesingle25 & -36.354\% & 2.331 & -38.087\% & 1.899 & \textbf{0.14} (0.53\%) \\
		& \underline{Ours ($C'$=4, $r$=32)} & -- & \underline{-56.337\%} & \underline{3.784} & -52.609\% & 3.305 & \underline{0.40} (1.50\%) \\
		& \textbf{Ours} & -- & \textbf{-61.578\%} & \textbf{4.153} & \textbf{-63.607\%} & \textbf{3.826} & 0.74 (2.78\%) \\
		\bottomrule
	\end{tabular}
\end{table*}

\section{Detailed Ablation Studies}
\label{sec:supDetailedAblationStudies}

\subsection{Detailed Ablation on Adapt-ICMH and SCA}
\label{subsec:supDetailedAblationAdaptICMHSCA}

To further investigate whether the performance gains of S$^2$-CoT stem from parameter capacity or architectural design, we provide an extended comparison with scaled variants of Adapt-ICMH \cite{Li2024}. As shown in \cref{ablation_sfma_sca}, our method (d) outperforms all Adapt-ICMH variants across both tasks. Notably, even when Adapt-ICMH is scaled to a higher parameter count (\eg, 0.64M in method (e)), it still falls short of S$^2$-CoT (0.42M). Conversely, while the lightweight Adapt-ICMH-64 (g) reduces overhead, its performance gap widens significantly. These results confirm that the efficacy of S$^2$-CoT is not a byproduct of parameter scaling. Instead, the gains are fundamentally driven by our SFA design, particularly the dynamic soft fusion mechanism and the role-matched synergy between SFA and SCA, which together confirm that structural optimization is more critical for performance than merely expanding the parameter budget.
\begin{table*}[!htbp]
	\caption{Ablation study of Adapt-ICMH \cite{Li2024} and SCA synergy (extension of Tab. 5 in the main paper).}
	\label{ablation_sfma_sca}
	\centering
	\small
	\begin{tabular}{ccccccccccc}
		\toprule
		\multirow[c]{2}{*}{Method} & \multirow[c]{2}{*}{SFA} & \multicolumn{2}{c}{SCA} & \multicolumn{2}{c}{Object Detection} & \multicolumn{2}{c}{Instance Segmentation} & \multirow[c]{2}{*}{\shortstack{Trainable\\Params$\downarrow$ (M)}} \\
		\cline{3-4}
		& &  $h_a$  &  $h_s$  & BD-Rate$\downarrow$ & BD-mAP$\uparrow$ & BD-Rate$\downarrow$ & BD-mAP$\uparrow$ & \\
		\midrule
		\textbf{(d) Ours} & $\checkmark$ & $\checkmark$ & $\checkmark$ &  \textbf{-60.824\%} & \textbf{4.014} & \textbf{-61.784\%} & \textbf{3.480} & \underline{0.42} (5.59\%) \\
		(e) & Adapt-ICMH-128 & $\checkmark$ & $\checkmark$ & -58.399\% & 3.709 & -60.936\% & 3.413 & 0.64(8.52\%) \\		
		(f) & Adapt-ICMH-96 & $\checkmark$ & $\checkmark$ & -59.811\% & 3.665 & -59.309\% & 3.341 & 0.48(6.39\%) \\
		(g) & Adapt-ICMH-64 & $\checkmark$ & $\checkmark$ & -57.894\% & 3.894 & -57.851\% & 3.356 & \textbf{0.30} (3.99\%) \\
		\bottomrule
	\end{tabular}
\end{table*}

\begin{figure}[!htbp]
	\centering
	\includegraphics[width=\columnwidth]{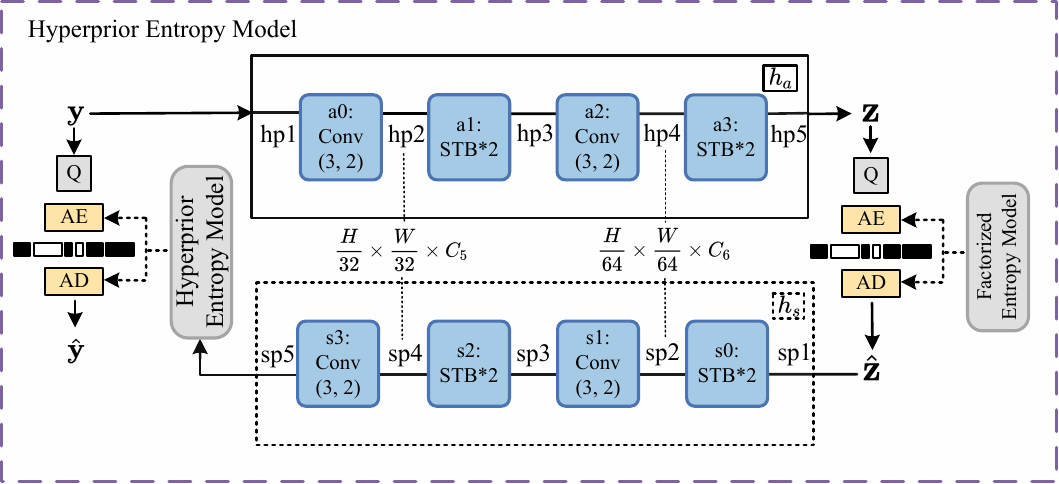} % Reduce the figure size so that it is slightly narrower than the column. Don't use precise values for figure width.This setup will avoid overfull boxes.
	\caption{Architecture of the hyperprior entropy model with location markers. The figure illustrates the composition of the hyper-encoder $h_a$ and hyper-decoder $h_s$.}
	\label{caa_pos}
\end{figure}
\subsection{Detailed Ablation Study on SCA Placement}
\label{subsec:supDetailedAblationStudyonSCAPlacement}

To determine the optimal placement of the SCA within the hyperprior entropy model, we conducted a series of ablation studies trained for 18 epochs. We designed multiple experimental schemes, where each scheme involves inserting SCA at different stages of the hyper-encoder $h_a$ and hyper-decoder $h_s$. The specific insertion placements for each scheme correspond to the positions labeled hp1–hp5 and sp1–sp5 in the architecture shown in \cref{caa_pos}. To quantify the performance of each placement, we measured the gain in mAP relative to the Adapt-ICMH  \cite{Li2024}. Specifically, for each experimental run, we computed the predicted mAP of Adapt-ICMH at the same bitrate using a fitted polynomial Rate-mAP curve. The final gain, $ \Delta\text{mAP} $, is then calculated as the difference between our method's achieved mAP and this predicted mAP value. A higher positive value indicates a more significant performance improvement attributable to the SCA's placement.

The experimental results are summarized in \cref{tab_4_caa_pos}. Although placing the adapter in the reverse bottleneck (hp1, sp5) or asymmetric layers (hp3, hp5, sp3, sp5) yields promising performance, these configurations typically come with increased parameter overhead. In contrast, inserting adapters into symmetric middle layers achieves a better trade-off between efficiency and performance, demonstrating superior overall effectiveness. This suggests that for the LIC architecture, the features at the intermediate middle spatial resolution retain a more optimal balance of semantic information and channel detail for the adapter to effectively operate on. 

The most compressed high layer features at the classic bottleneck, while semantically rich, may have discarded spatial and channel cues that are crucial for the adapter's refinement process in the model. The superior performance at hp3 and sp4, coupled with its minimal parameter footprint, validates the selection of the symmetric middle layer insertion strategy for our final architecture.

\begin{table}[!h]
	\caption{Ablation study on the placement of SCA within the entropy model. Gain ($ \Delta\text{mAP} $) is measured as the difference in mAP between our method's result and the value predicted by the Adapt-ICMH Rate-mAP curve at the same bitrate.}
	\label{tab_4_caa_pos}
	\centering
	\small
	\setlength{\tabcolsep}{0.6mm}
	\begin{tabular}{cccc}
		\toprule
		Placement in $h_a, h_s$ & Core Idea & $ \Delta\text{mAP} $ $\uparrow$ & Params$\downarrow$ \\
		\midrule
		hp1, sp5 & reverse bottleneck & +0.459 & 0.497 \\
		\midrule
		hp5, sp1 & classic bottleneck & -0.496 & 0.421 \\
		\midrule
		\multirow{2}{*}{\shortstack{hp3, sp4 or\\hp3, sp3}} & \multirow{2}{*}{\shortstack{symmetrical\\middle layer}} & +0.479 & 0.421 \\
		&&&\\
		\midrule
		\multirow{2}{*}{\shortstack{hp2, hp4,\\sp2, sp4}}  & \multirow{2}{*}{\shortstack{symmetrical middle\\high layer}} & +0.242 & 0.438 \\
		&&&\\
		\midrule
		\multirow{2}{*}{\shortstack{hp3, hp5,\\sp1, sp3}} & \multirow{2}{*}{\shortstack{symmetrical middle\\high layer}} & +0.343 & 0.438 \\
		&&&\\
		\midrule
		hp3, hp5, sp3, sp5 & asymmetric layer & +0.467 & 0.504 \\
		\bottomrule
	\end{tabular}
\end{table}

\subsection{Detailed Hyperparameters Analysis}
\label{subsec:supDetailedHyperparametersAnalysis}

This subsection supplements the main paper by providing a comprehensive analysis of the key hyperparameters governing our adapters' capacity: the Structural Fidelity Adapter's (SFA) middle dimension, $C'$, and the Semantic Context Adapter's (SCA) reduction ratio, $ r $. $C'$ controls the SFA's representational power for structural features, while $ r $ adjusts the SCA's bottleneck for context modeling.
\begin{table}[!htbp]
	\caption{Comprehensive ablation study on the SFA middle dimension $C'$ and the SCA reduction ratio $r$.}
	\label{tab_3_c_r_ablation}
	\centering
	\small
	\setlength{\tabcolsep}{0.5mm}
	\begin{tabular}{ccccccc}
		\toprule
		\multirow{2}{*}{$C'$} & \multirow{2}{*}{$ r $} 
		& \multicolumn{2}{c}{Detection} 
		& \multicolumn{2}{c}{Segmentation} 
		& \multirow[c]{2}{*}{\shortstack{Params\\$\downarrow$(M)}} \\
		& & BD-Rate$\downarrow$ & BD-mAP$\uparrow$ & BD-Rate$\downarrow$ & BD-mAP$\uparrow$ & \\
		\midrule
		32 & 32 & -60.276\% & 3.852 & -60.255\% & 3.386 & 0.28 (3.73\%) \\
		32 & 16 & -57.622\% & 3.875 & -61.991\% & 3.462 & 0.29 (3.86\%) \\
		64 & 8  & -60.824\% & 4.014 & -61.784\% & 3.480 & 0.42 (5.59\%) \\
		96 & 4  & -61.329\% & 4.063 & -62.429\% & 3.529 & 0.57 (7.59\%) \\
		128 & 4 & -62.763\% & 4.165 & -63.657\% & 3.625 & 0.72 (9.59\%) \\
		\bottomrule
	\end{tabular}
\end{table}

\begin{table}[htbp]
	\caption{Performance comparison. Top: compression ratio and mAP on original images. Bottom: performance on three detectors.}
	\centering
	\small
	\setlength{\tabcolsep}{0.2pt}
	\begin{tabular}{ccccc}
		\toprule
		& \multirow{2}{*}{Original Image} & \multicolumn{2}{c}{LIC} & Traditional \\
		\cmidrule(lr){3-4} \cmidrule(lr){5-5}
		&  & Base Codec & \textbf{Ours} & JPEG(Q10) \\
		\midrule
		Compression Ratio & None & 219:1 & \textbf{296:1} & 67:1 \\
		mAP$\uparrow$   & 51.302 & 45.231 & \textbf{47.760} & 15.481 \\
		\bottomrule
		\toprule
		\multirow{2}{*}{Methods} & \multicolumn{2}{c}{Base Codec} & \multicolumn{2}{c}{\textbf{Ours}} \\
		\cmidrule(lr){2-3} \cmidrule(lr){4-5}
		& bpp$\downarrow$ & mAP$\uparrow$ & bpp$\downarrow$ & mAP$\uparrow$ \\
		\midrule
		Faster-RCNN & 0.1096 & 31.606 & \textbf{0.0812} & \textbf{36.601} \\
		YOLOv11m    & 0.1096 & 45.231 & \textbf{0.0812} & \textbf{47.760} \\
		YOLO26x     & 0.1096 & 49.972 & \textbf{0.0812} & \textbf{52.410} \\
		\bottomrule
	\end{tabular}
	\label{yolo_comparison}
\end{table}
The complete results of the controlled ablation study are presented in \cref{tab_3_c_r_ablation}. The data reveals a clear trade-off between performance and parameter complexity across various settings. The configuration selected for our main experiments ($C'$=64, $ r $=8) was determined from this analysis to provide the optimal balance between task accuracy and efficiency, noting their coupled influence.

\begin{table*}[h]
	\caption{More Detailed Comparison on computational complexity.}
	\label{more_complexity_comparison_compact}
	\centering
	\small
	\setlength{\tabcolsep}{1mm}
	\begin{tabular*}{\textwidth}{@{\extracolsep{\fill}} c cccccccc}
		\toprule
		\multirow{2}{*}{Model} 
		& \multicolumn{2}{c}{KMACs/pixel} 
		& \multicolumn{2}{c}{CPU Average Latency (ms)} 
		& \multicolumn{2}{c}{GPU Average Latency (ms)} 
		& \multirow{2}{*}{\shortstack{Trainable\\Params $\downarrow$ (M)}}
		& \multirow{2}{*}{\shortstack{BD-\\Acc $\uparrow$}} \\
		
		\cmidrule(lr){2-3} \cmidrule(lr){4-5} \cmidrule(lr){6-7} 
		
		& \hspace{15pt}Enc.\hspace{15pt} & \hspace{15pt}Dec.\hspace{15pt} & \hspace{15pt}Enc.\hspace{15pt} & \hspace{15pt}Dec.\hspace{15pt} & \hspace{15pt}Enc.\hspace{15pt} & Dec. & & \\
		\midrule
		base codec & 132.64 & 176.42 & 94.87 & 94.75 & 33.22 & 31.18 & - & - \\  \midrule
		full fine-tuning & 132.64 & 176.42 & 94.87 & 94.75 & 33.22 & 31.18 & 7.51 & 17.7 \\
		Adapt-ICMH\cite{Li2024} & 147.39 & 191.16 & 111.01 & 110.14 & 29.49 & 25.34 & 0.29 & 16.9 \\
		Ours & 153.57 & 197.34 & 109.49 & 101.03 & 29.60 & 28.45 & 0.42 & 17.4 \\
		\bottomrule
	\end{tabular*}
\end{table*}
\section{Generalization to YOLO Detectors}
\label{sec:supGeneralizationYOLODetectors}

At a compression ratio of 296:1, S$^2$-CoT maintains robust performance (\eg, 47.760 mAP on YOLOv11m), narrowing the gap to the unconstrained original image upper bound presented in \cref{yolo_comparison}. Meanwhile, we also evaluate S$^2$-CoT across multiple downstream detectors, including Faster-RCNN, YOLOv11m \cite{Khanam2024}, and YOLO26x \cite{Sapkota2025}. The results demonstrate consistent mAP gains while reducing the bpp from 0.1096 to 0.0812, confirming that our method provides a detector-agnostic and highly efficient solution for machine vision compression.

\section{Computational Complexity and Efficiency}
\label{sec:supComputationalComplexityAndEfficiency}
Complementing the efficiency analysis in the main paper, we provide a comprehensive breakdown of computational complexity and inference latency in \cref{more_complexity_comparison_compact}. Acknowledging the hardware-dependent nature of runtime performance, we evaluate the average encoding and decoding latency on both an INTEL(R) XEON(R) SILVER 4510 CPU and an NVIDIA L40 GPU. The results verify that our proposed framework maintains competitive inference speeds with negligible computational overhead (in KMACs/pixel) compared to full fine-tuning, validating its practical efficiency alongside superior parameter reduction.

\section{Integration Details of Adapters}
\label{sec:supIntegrationDetailsOfAdapters}

To demonstrate the architectural universality and seamless plug-and-play capability of our proposed framework, we detail the integration of SFA and SCA into four distinct, state-of-the-art learned image compression baselines. These codecs represent a diverse spectrum of architectural paradigms, encompassing both Transformer-based models ($ \textit{Lu2022-TIC} $ \cite{Lu2021} and $ \textit{DCAE} $ \cite{Lu2025}) and CNN-based models ($ \textit{Cheng2020-anchor} $ \cite{Cheng2020} and $ \textit{ELIC} $ \cite{He2022a}). Crucially, they feature highly heterogeneous entropy modeling mechanisms, ranging from standard hyperpriors and spatial autoregression to sophisticated dictionary-based cross-attention and channel-conditional backward-adaptive models, each tailored to distinct compression scenarios.

\begin{figure*}[!htbp]
	\centering
	%	, height=3.7cm
	\includegraphics[width=2\columnwidth, height=3.7cm]{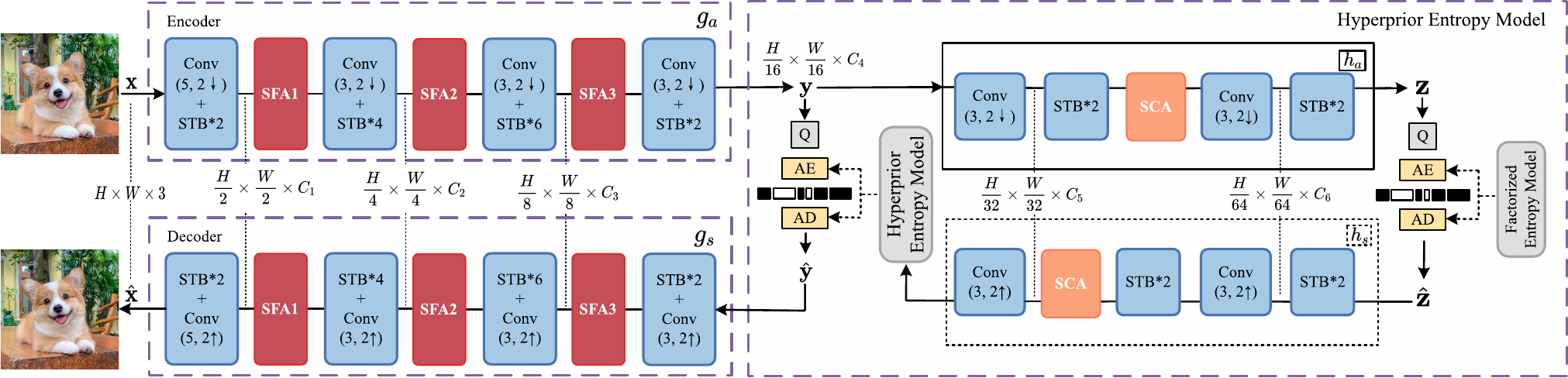} % Reduce the figure size so that it is slightly narrower than the column. Don't use precise values for figure width.This setup will avoid overfull boxes.
	\caption{Our SFA and SCA are integrated into the $ \textit{Lu2022-TIC} $ codec \cite{Lu2021}. STB denotes the Swin-Transformer Block. Conv(n, 2↑) denotes a transposed convolution (kernel size=n) with a stride of 2 for upsampling, and Conv(n, 2↓) denotes a convolution (kernel size=n) with a stride of 2 for downsampling.}
	\label{fig_tic}
\end{figure*}

\begin{figure*}[!htbp]
	\centering
	\includegraphics[width=2\columnwidth, height=3.7cm]{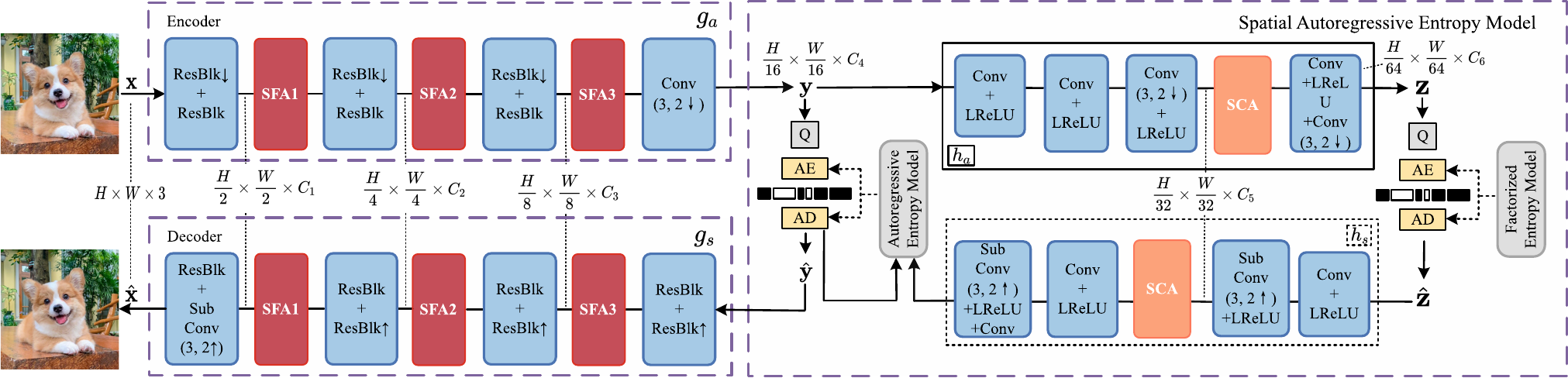} % Reduce the figure size so that it is slightly narrower than the column. Don't use precise values for figure width.This setup will avoid overfull boxes.
	\caption{Our SFA and SCA are integrated into the $ \textit{Cheng2020-anchor} $ codec \cite{Cheng2020}. ResBlk denotes a residual block, where ResBlk↓ indicates a downsampling residual block with a stride of 2. Conv represents a standard $3 \times 3$  convolution. LReLU denotes Leaky ReLU. Sub Conv(3, 2↑) denotes a sub-pixel convolution (kernel size=3) with a stride of 2 for upsampling.}
	\label{fig_cheng2020}
\end{figure*}

\begin{figure*}[!htbp]
	\centering
	\includegraphics[width=2\columnwidth, height=3.8cm]{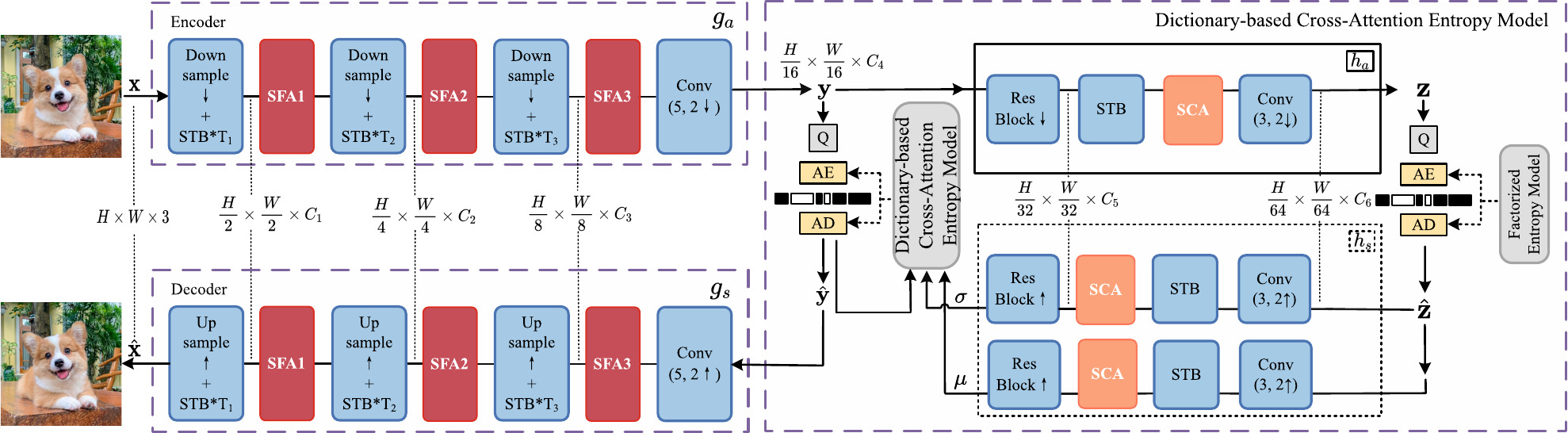} % Reduce the figure size so that it is slightly narrower than the column. Don't use precise values for figure width.This setup will avoid overfull boxes.
	\caption{Our SFA and SCA are integrated into the $ \textit{DCAE} $ codec \cite{Lu2025}. Downsample denotes the ResidualBottleneckBlockWithStride blocks, where STB indicates the SwinBlockWithConvMulti blocks. Conv(5, 2↓) represents a $5 \times 5$  convolution with a stride of 2. Upsample denotes the ResidualBottleneckBlockWithUpsample blocks. Conv(5, 2↑) represents a $5 \times 5$  convolution with a stride of 2 for upsampling.}
	\label{fig_dcae}
\end{figure*}

\begin{figure*}[!htbp]
	\centering
	\includegraphics[width=2\columnwidth, height=3.7cm]{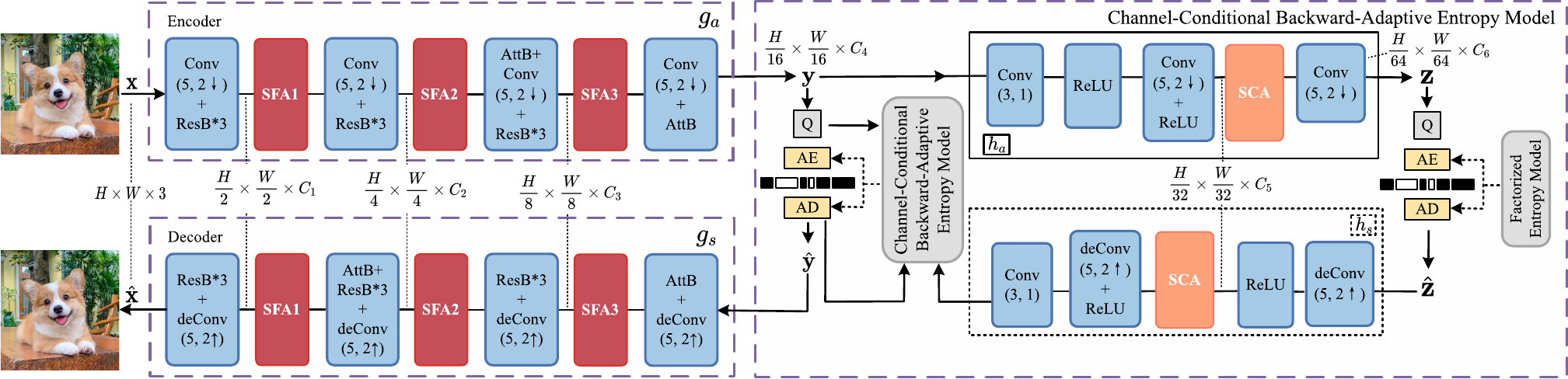} % Reduce the figure size so that it is slightly narrower than the column. Don't use precise values for figure width.This setup will avoid overfull boxes.
	\caption{Our SFA and SCA are integrated into the $ \textit{ELIC} $ codec \cite{He2022a}. Conv(5, 2↓) represents a $5 \times 5$  convolution with a stride of 2. ResB denotes residual blocks, and AttB denotes attention blocks.. Conv(3, 1) represents a standard $3 \times 3$  convolution with a padding of 1. The deConv(5, 2↑) denotes a $5 \times 5$ transposed convolution with a stride of 2, serving as a learnable spatial upsampling operation.}
	\label{fig_elic}
\end{figure*}

As illustrated in \cref{fig_tic}, \cref{fig_cheng2020}, \cref{fig_dcae}, and \cref{fig_elic}, our dual-adapter framework maintains a consistent integration strategy despite the structural variations of the base codecs. The SFA modules are strategically interleaved within the multiscale stages of the encoder-decoder backbones ($g_a, g_s$) to preserve high-fidelity structural information in the feature domain. Concurrently, the SCA modules are embedded within the symmetrical intermediate layers of the entropy models ($h_a, h_s$). This placement allows the SCA to refine the semantic context effectively, regardless of whether the underlying probability estimation relies on local spatial neighbors, channel contexts, or global dictionary priors. This uniform applicability across diverse backbones and statistical models strongly validates the robustness of our structure-semantics synergy.

\section{More Results on the Diverse Base Codec}
\label{sec:supMoreResultsOnTheDiverseBaseCodec}

\begin{table*}[!htbp]
	\caption{Quantitative comparison for object detection and instance segmentation tasks based on the $ \textit{DCAE} $ and $ \textit{ELIC} $ codec. Our method is competitive with full fine-tuning against others. Our method demonstrates superior BD-metric performance and parameter efficiency.}
	\label{tic_results}
	\centering
	\small
	\setlength{\tabcolsep}{8.6pt}
	\begin{tabular}{cccccccc}
		\toprule
		\multirow[c]{2}{*}{Base} & \multirow[c]{2}{*}{Method} & \multirow[c]{2}{*}{Venue} & \multicolumn{2}{c}{Object Detection} & \multicolumn{2}{c}{Instance Segmentation} & \multirow[c]{2}{*}{\shortstack{Trainable\\Params$\downarrow$ (M)}} \\
		%		\cline{4-7}
		& & & BD-Rate$\downarrow$ & BD-mAP$\uparrow$ & BD-Rate$\downarrow$ & BD-mAP$\uparrow$ & \\
		\midrule
		\multirow[c]{6}{*}{$ \textit{DCAE} $}
		& full fine-tuning & -- & -84.677\% & 4.002 & -81.897\% & 3.049 & 119.40 (100.00\%) \\ %\cline{2-8}
		& Channel Selection & ICPR\textquotesingle22 & -24.498\% & 0.977 & -18.661\% & 0.747 & 2.68 (2.24\%) \\
		& TransTIC & ICCV\textquotesingle23 & -49.269\% & 2.339 & -50.120\% & 2.271 & 1.13 (0.95\%) \\
		& Adapt-ICMH & ECCV\textquotesingle24 & -64.608\% & 2.863 & -63.240\% & 2.404 & \underline{0.36} (0.30\%) \\
		& SVD-LoRA & CVPR\textquotesingle25 & -58.166\% & 2.300 & -54.371\% & 2.229 & \textbf{0.17} (0.14\%) \\
		& \textbf{Ours} & -- & \textbf{-68.882\%} & \textbf{3.361} & \textbf{-68.363\%} & \textbf{2.743} & 0.42 (0.35\%) \\
		\midrule
		\multirow[c]{6}{*}{ $ \textit{ELIC} $ }
		& full fine-tuning & -- & -70.409\% & 5.745 & -70.153\% & 4.158 & 33.79 (100.00\%) \\ 
		& Channel Selection & ICPR\textquotesingle22 & -23.235\% & 1.518 & -11.041\% & 0.591 & 2.68 (7.93\%) \\
		& Adapt-ICMH & ECCV\textquotesingle24 & -54.844\% & 3.379 & -51.173\% & 2.861 & \textbf{0.41} (1.21\%) \\
		& SVD-LoRA & CVPR\textquotesingle25 & -41.190\% & 2.149 & -42.721\% & 1.948 & \textbf{0.41} (1.21\%) \\
		& \textbf{Ours} & -- & \textbf{-60.046\%} & \textbf{4.051} & \textbf{-59.214\%} & \textbf{3.174} & \underline{0.74} (2.19\%) \\		
		\bottomrule
	\end{tabular}
\end{table*}
\begin{figure*}[!htbp]
	\centering
	\begin{tabular}{c:cc}
		% Row 1: Three images
		\includegraphics[width=0.31\textwidth]{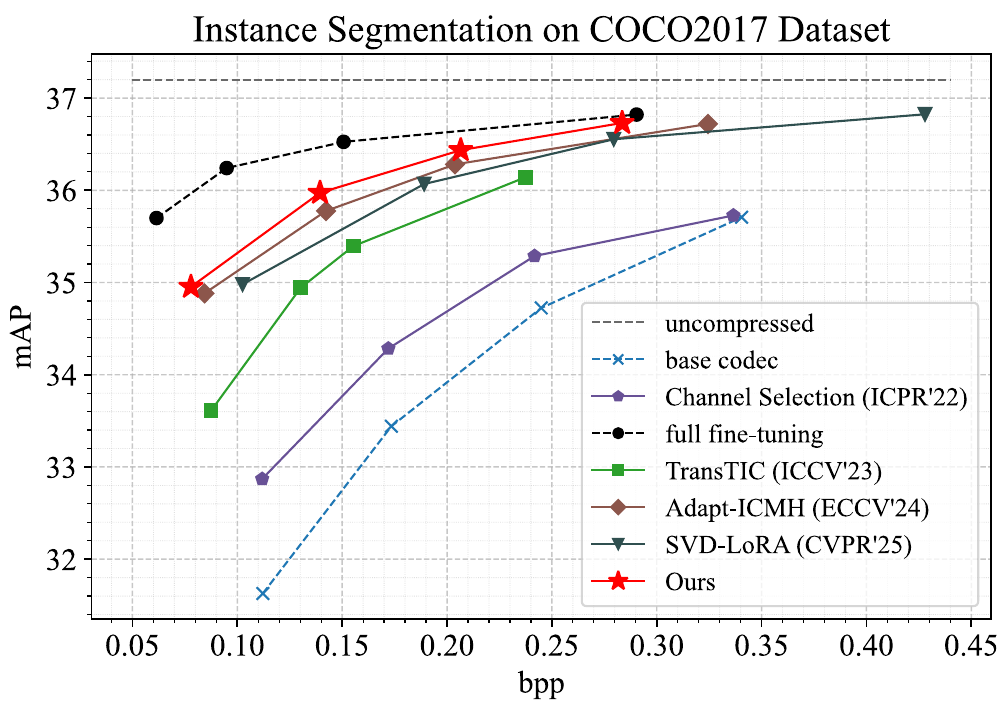} &
		\includegraphics[width=0.31\textwidth]{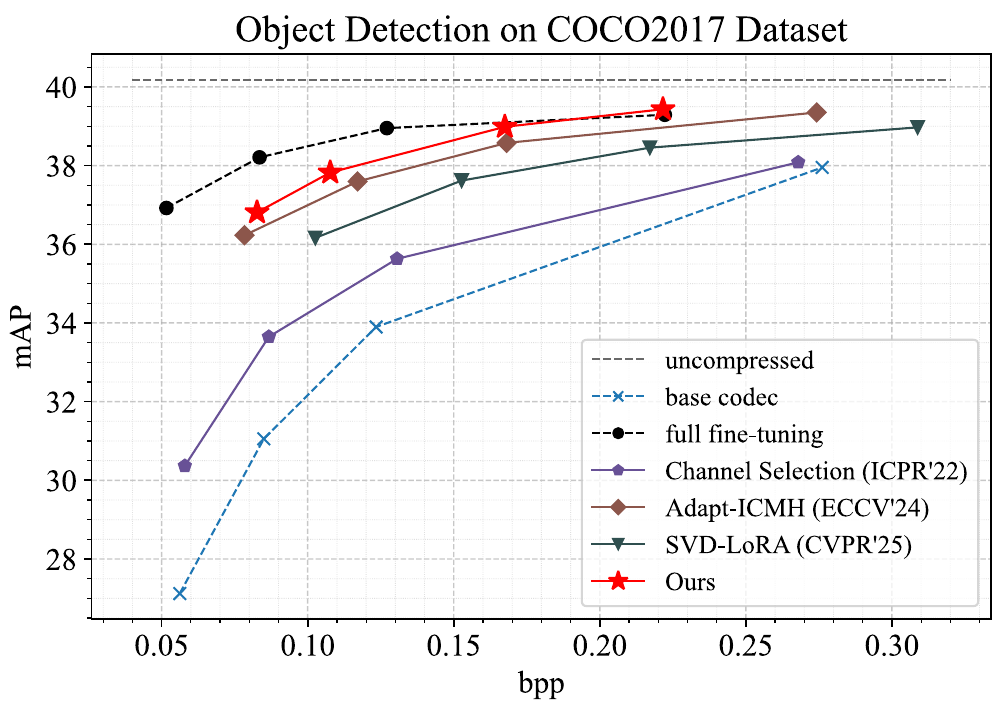} &
		\includegraphics[width=0.31\textwidth]{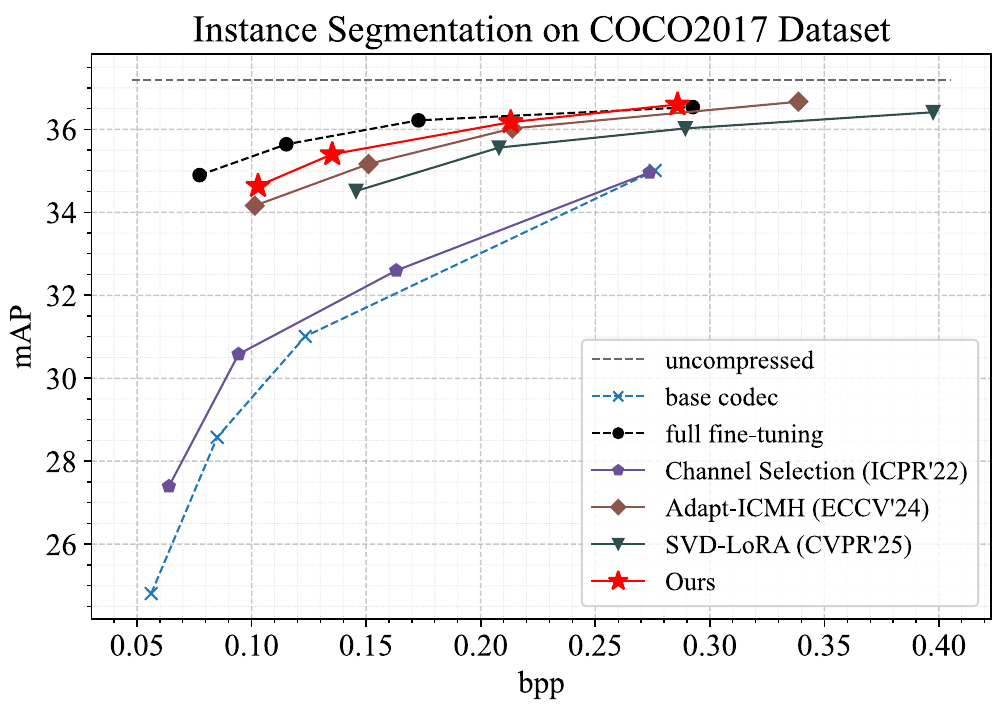} \\
		% Row 2: Empty merged cell, and the (b) label for the image above
		(a) & \multicolumn{2}{c}{(b)} \\
	\end{tabular}
	\caption{Comparison of rate-accuracy performance across various tasks and base codecs. (a) The instance segmentation results on the $ \textit{DCAE} $ base codec. (b) Object detection (left) and instance segmentation (right) performance on the $ \textit{ELIC} $ base codec.}
	\label{appendix_comparison}
\end{figure*}

To further substantiate the scalability and architectural universality of our S$^2$-CoT framework, we extend our evaluation to two advanced, high-performance base codecs: $ \textit{DCAE} $ \cite{Lu2025}, a large-scale Transformer-based model utilizing dictionary-based cross-attention, and $ \textit{ELIC} $ \cite{He2022a}, a sophisticated CNN-based model featuring unevenly grouped space-channel contextual coding. As presented in \cref{tic_results}, our method demonstrates exceptional robustness when applied to these complex architectures. Specifically, on the $ \textit{DCAE} $ codec, S$^2$-CoT outperforms existing PEFT methods, such as TransTIC and Adapt-ICMH, across both object detection and instance segmentation tasks. Notably, as visualized in \cref{appendix_comparison} (a), our method establishes a rate-accuracy frontier that closely parallels the theoretical upper bound of full fine-tuning, all while adding only a negligible fraction of trainable parameters relative to the large-scale backbone.

Similarly, the experiments on the $ \textit{ELIC} $ codec validate the superior efficacy of our structure-semantics synergy. As shown in the bottom section of \cref{tic_results} and \cref{appendix_comparison} (b), S$^2$-CoT consistently dominates the performance landscape, delivering superior coding efficiency and detection accuracy compared to recent state-of-the-art adaptations like SVD-LoRA and Adapt-ICMH. These results collectively confirm that our synergistic adaptation strategy is model-agnostic, effectively mitigating the domain gap for machine vision tasks even when integrated into highly optimized, state-of-the-art compression baselines.

\section{Quantitative Results on Classification Task}
\label{sec:supQuantitativeResultsOnClassificationTask}

This section provides detailed quantitative data on classification task for the generalization experiment on the $ \textit{Lu2022-TIC} $ codec, supplementing the rate-accuracy curves in the main paper.
The BD-metrics in \cref{tab_cls} quantify the performance of different Parameter-Efficient Fine-Tuning
\begin{table}[!htbp]
	\caption{Quantitative comparison of PEFT methods on the $ \textit{Lu2022-TIC} $ codec for the classification task. Our method demonstrates superior BD-metric performance and parameter efficiency.}
	\label{tab_cls}
	\centering
	\small
	\begin{tabular}{cccc}
		\toprule
		\multirow{2}{*}{Method} & \multicolumn{2}{c}{Classification} & \multirow{1}{*}{Params} \\ 
		& BD-Rate$\downarrow$ & BD-Acc$\uparrow$ & $\downarrow$(M)\\
		\midrule
		full fine-tuning  & - & 17.688 & 7.51 (100\%) \\
		Channel Selection & -37.178\% & 6.278 & 0.92 (12.25\%) \\
		TransTIC & -58.529\% & 9.956 & 1.62 (21.57\%) \\
		ICMH-Net & -18.759\% & 3.360 & 3.98 (53.00\%) \\
		Adapt-ICMH & -88.573\% & 16.901 & 0.29 (3.86\%) \\
		SVD-LoRA & -50.162\% & 7.920 & 0.09 (1.20\%) \\
		Ours & \textbf{-93.453\%} & \textbf{17.351} & 0.42 (5.59\%) \\
		\bottomrule
	\end{tabular}
\end{table}
(PEFT) ways. The data confirms that our method significantly outperforms other PEFT approaches in both BD-Rate savings and BD-Acc gains. This validates the broad applicability of our method, showing that the synergistic strategy can be successfully extended to varying tasks beyond object detection, such as classification, without compromising efficiency.

\section{Sensitivity \& Necessity of Co-Tuning}
\label{sec:supSensitivityNecessityOfCo-Tuning}

We investigate deploying SCA in isolation, and \cref{sensitivity_analysis} shows its solitary application causes counter-intuitive performance deterioration,
\begin{table}[h]
	\centering
	\caption{Isolated SCA reveals the entropy model’s inherent hypersensitivity, where even minimal parameter addition leads to performance collapse, while our S$^2$-CoT reverses this trend.}
	\label{sensitivity_analysis}
	\small
	\setlength{\tabcolsep}{6.5pt}
	\begin{tabular}{lccc}
		\toprule
		Method & \multicolumn{2}{c}{Object Detection} & Trainable \\
		$ \textit{Lu2022-TIC} $ & BD-Rate $\downarrow$ & BD-mAP $\uparrow$ & Params $\downarrow$ (M) \\
		\midrule
		+SFA & -58.077\% & 3.842 & 0.40 (5.33\%) \\
		+SCA & +8.160\% & -0.511 & 0.02 (0.27\%) \\
		+SFA +SCA & -60.824\% & 4.014 & 0.42 (5.59\%) \\
		\bottomrule
	\end{tabular}
\end{table}
highlighting the entropy model’s extreme hypersensitivity to uncoordinated adjustments. By synchronizing structural and statistical adaptation, our S$^2$-CoT framework converts this degradation into synergistic gains, achieving SOTA across diverse base codecs.

\section{Framework Paradigm and Modularity}
\label{sec:supFrameworkParadigmAndModularity}

The proposed S$^2$-CoT framework introduces two types of lightweight adapters to efficiently regulate key components of the transform module and the entropy model. Specifically, the SFA harmonizes spatial and frequency representations for flexible feature modulation, while the SCA focuses on calibrating channel-wise statistics in the entropy model to facilitate precise probability estimation. From an architectural perspective, S$^2$-CoT establishes a structured, efficient fine-tuning scheme. Benefiting from a core design that decouples the regulation mechanism from the backbone network, the proposed adapters essentially function as interchangeable plug-ins. This implies that the framework is not dependent on specific component implementations: the modules used to regulate these two key structures can be substituted by analogous architectures, rather than being confined to the designs presented herein. Simultaneously, this modularity enables S$^2$-CoT to be seamlessly integrated into diverse LIC architectures.

\section{PyTorch Implementation of SFA and SCA}
\label{sec:supPyTorchImplementationOfSFAAndSCA}

To provide a deeper understanding of our S$^2$-CoT framework, we detail the PyTorch implementation of its two core components: SFA and SCA. These adapters, designed to achieve a crucial structure–semantics synergy, are implemented as lightweight and plug-and-play network blocks. The full, runnable source code for the S$^2$-CoT can be found in the code folder of the supplementary material.

\begin{lstlisting}
	
	class SFA(nn.Module):
	"""
	Spatial-Frequency Structural Fidelity Adapter: Feature Enhancement via Three Stages:
	1. Channel Excitation and Bottleneck Projection.
	2. Spatial-Frequency Dual-Branch Modulation.
	3. Soft Fusion.
	"""
	
	def __init__(self, in_dim: int = 128, middle_dim: int = 64, r: int = 16, se_factor: float = 1.0, adapt_factor: float = 1.0):
	super().__init__()
	self.adapt_factor = adapt_factor
	
	# Channel
	self.c_squeeze = nn.AdaptiveAvgPool2d(1)
	self.c_excite = nn.Sequential(
	nn.Conv2d(in_dim, in_dim // r, 1, bias=False),
	nn.ReLU(),
	nn.Conv2d(in_dim // r, in_dim, 1, bias=False),
	nn.Sigmoid()
	)
	self.down1 = nn.Conv2d(in_dim, middle_dim, kernel_size=1)
	self.se_alpha = nn.Parameter(torch.tensor(se_factor))
	
	# Spatial branch
	self.s_gate = nn.Conv2d(in_dim, middle_dim, kernel_size=1)
	self.s_dw3 = nn.Conv2d(middle_dim, middle_dim, kernel_size=5, padding=2, groups=middle_dim)
	self.s_up_proj = nn.Conv2d(middle_dim, in_dim, kernel_size=1)
	
	# Frequency branch
	self.f_dw_conv = nn.Conv2d(middle_dim, middle_dim, kernel_size=3, padding=1, groups=middle_dim)
	self.amplitude_mlp = nn.Conv2d(middle_dim, middle_dim, kernel_size=1)
	self.f_up_proj = nn.Conv2d(middle_dim, in_dim, kernel_size=1)
	self.amplitude_gelu = nn.GELU()
	self.ifft_relu = nn.ReLU()
	self.sigmoid = nn.Sigmoid()
	
	# Shared
	self.sf_dw3 = nn.Conv2d(in_dim, in_dim, kernel_size=3, padding=1, groups=in_dim, bias=False)
	self.sf_relu = nn.ReLU()
	
	# Fusion
	self.sf_mlp = nn.Sequential(
	nn.Conv2d(in_dim * 2, in_dim // 2, kernel_size=1, bias=False),
	nn.ReLU(),
	nn.Conv2d(in_dim // 2, in_dim, kernel_size=1, bias=False)
	)
	
	def forward(self, x: torch.Tensor) -> torch.Tensor:
	_, _, H, W = x.shape
	
	# Channel
	c_weights = self.c_squeeze(x)
	c_weights = self.c_excite(c_weights)
	c = c_weights * x * self.se_factor
	down = self.down1(x + c)
	
	# Spatial branch
	gated = self.s_dw3(down) * self.s_gate(x)
	s = self.s_up_proj(torch.relu(gated))
	
	# Frequency branch
	fft = torch.fft.rfft2(down, dim=(2, 3), norm='backward')
	amplitude, phase = torch.abs(fft), torch.angle(fft)
	amplitude = self.amplitude_mlp
	(self.amplitude_gelu(
	self.f_dw_conv(amplitude)))
	modulated = amplitude * self.sigmoid(amplitude)
	complex = torch.complex(modulated * torch.cos(phase), modulated * torch.sin(phase))
	ifft = torch.fft.irfft2(complex, s=(H, W), norm='backward')
	f = self.f_up_proj(self.ifft_relu(ifft))
	
	# Shared
	s = self.sf_relu(self.sf_dw3(s))
	f = self.sf_relu(self.sf_dw3(f))
	
	# Fusion
	fused = self.sf_mlp(torch.cat([s, f], dim=1))
	
	return x + c + fused * self.adapt_factor
	
\end{lstlisting}

\begin{lstlisting}	
	
	class SCA(nn.Module):
	"""
	Semantic Context Adapter (SCA).
	"""
	
	def __init__(self, channels: int, reduction_ratio: int = 8, adapt_factor: float = 1.0):
	super().__init__()
	self.adapt_factor = adapt_factor
	hidden_dim = max(4, channels // reduction_ratio)
	self.refine = nn.Sequential(
	nn.Conv2d(channels, hidden_dim, 1, bias=False),
	nn.ReLU(),
	nn.Conv2d(hidden_dim, channels, 1, bias=False)
	)
	self.se_inter = nn.Sequential(
	nn.AdaptiveAvgPool2d(1),
	nn.Conv2d(channels, hidden_dim, 1),
	nn.ReLU(),
	nn.Conv2d(hidden_dim, channels, 1),
	nn.Sigmoid()
	)
	
	def forward(self, x: torch.Tensor) -> torch.Tensor:
	refine = x + self.refine(x)
	se = self.se_inter(refine)
	return se * refine * self.adapt_factor
\end{lstlisting}

\section{Future Work}
\label{sec:supFutureWork}

The proposed S$^2$-CoT achieves state-of-the-art performance across multiple datasets and backbones, pushing the boundaries of task-oriented image compression with acceptable overhead. Given the constraints of computational resources and parameter size on practical deployment, our future work will focus on exploring lighter compression frameworks. We aim to optimize the trade-off between performance and efficiency by investigating diverse adapter structures and utilizing partial codec components. Furthermore, we plan to validate our method on a broader spectrum of downstream tasks and extend the current single-task adaptation paradigm to multi-task learning scenarios, thereby reducing training costs and enhancing generalization capabilities.

\section{More Qualitative Results}
\label{sec:supMoreQualitativeResultes}

\begin{figure*}[!htbp]
	\centering
	\includegraphics[width=1.8\columnwidth, height=11.5cm]{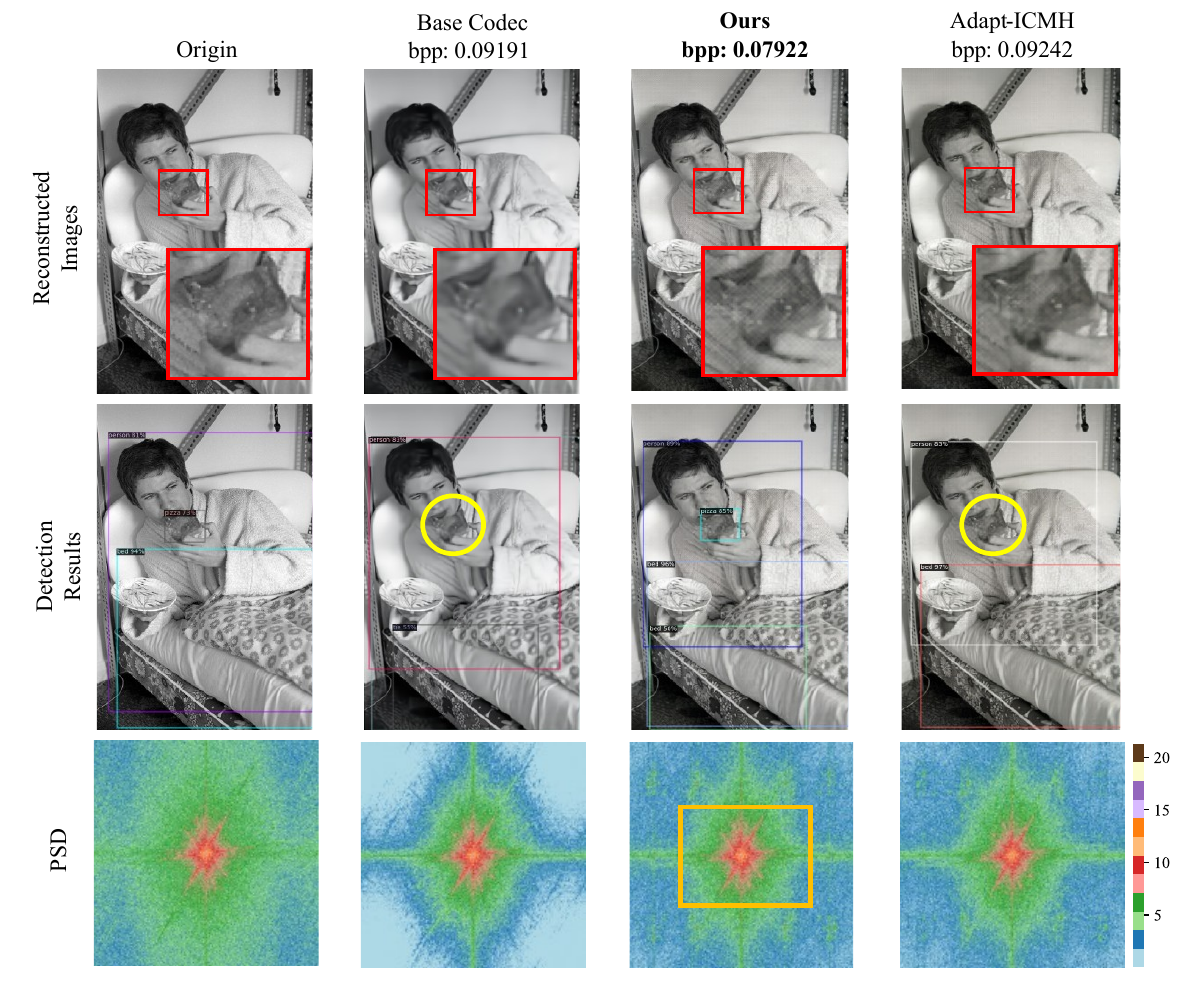} 
	\caption{More detection qualitative results.}
	\label{254368}
\end{figure*}

\begin{figure*}[!htbp]
	\centering
	\includegraphics[width=1.7\columnwidth, height=9cm]{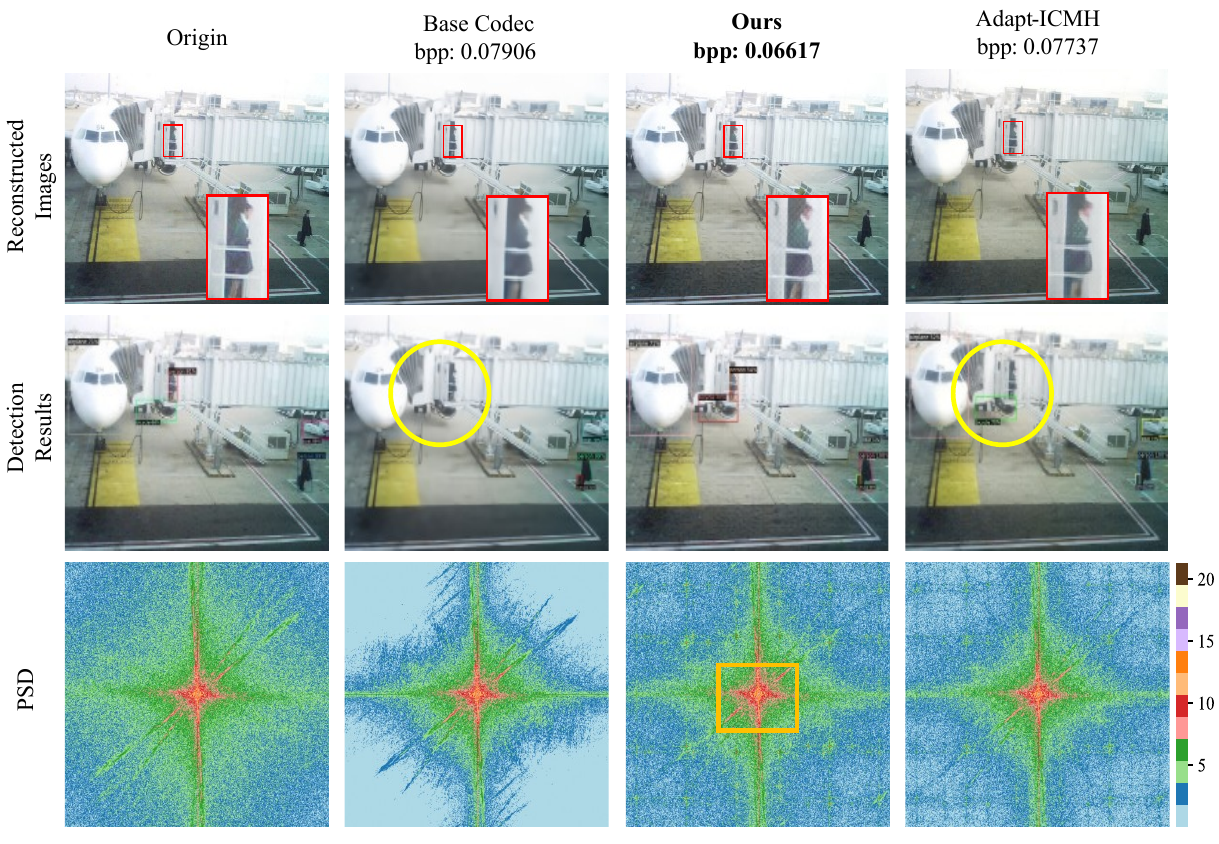} 
	\caption{More detection qualitative results.}
	\label{348881}
\end{figure*}

\begin{figure*}[!htbp]
	\centering
	\includegraphics[width=1.82\columnwidth, height=11cm]{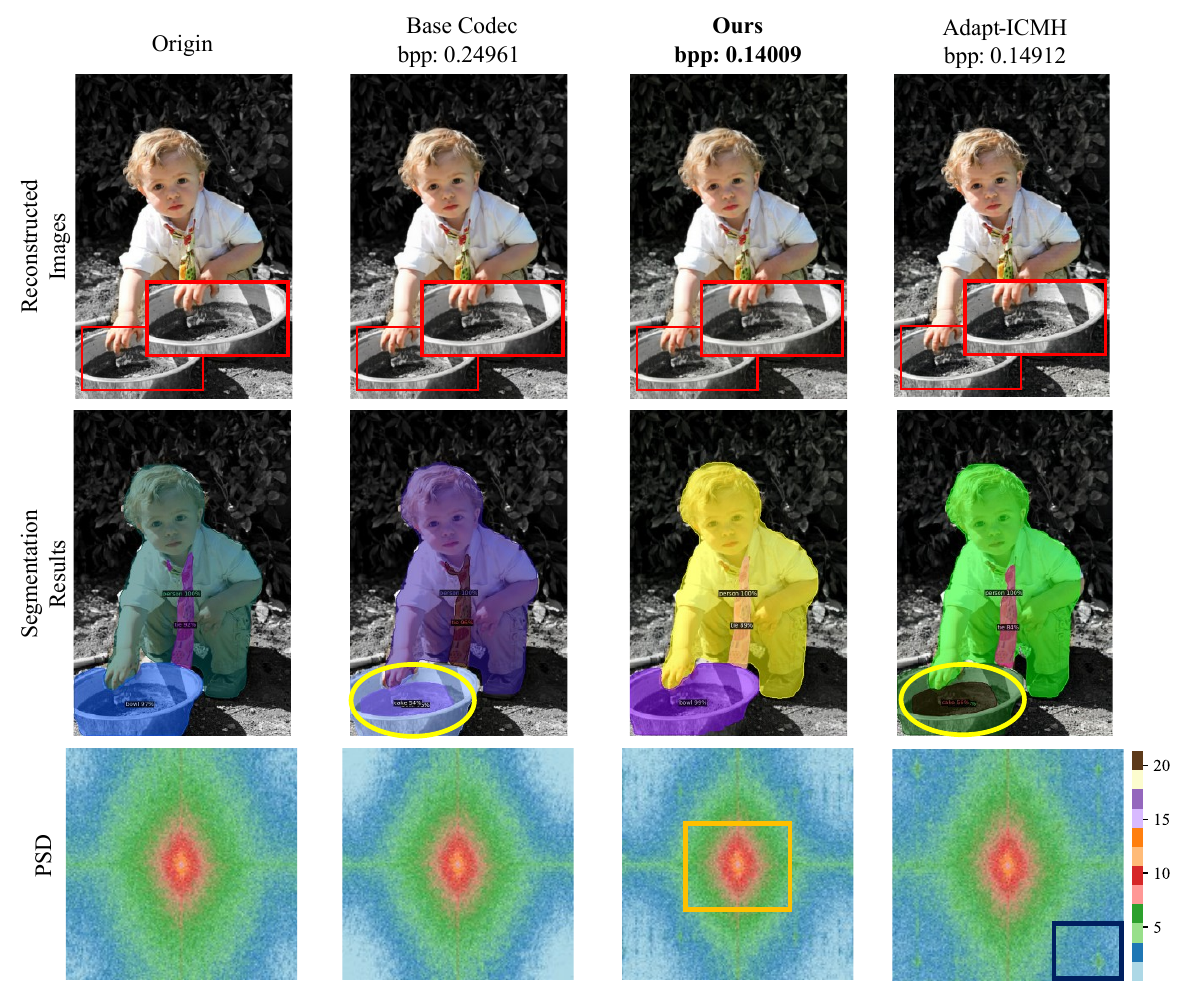} 
	\caption{More segmentation qualitative results.}
	\label{20333}
\end{figure*}

\begin{figure*}[!htbp]
	\centering
	\includegraphics[width=1.8\columnwidth, height=9cm]{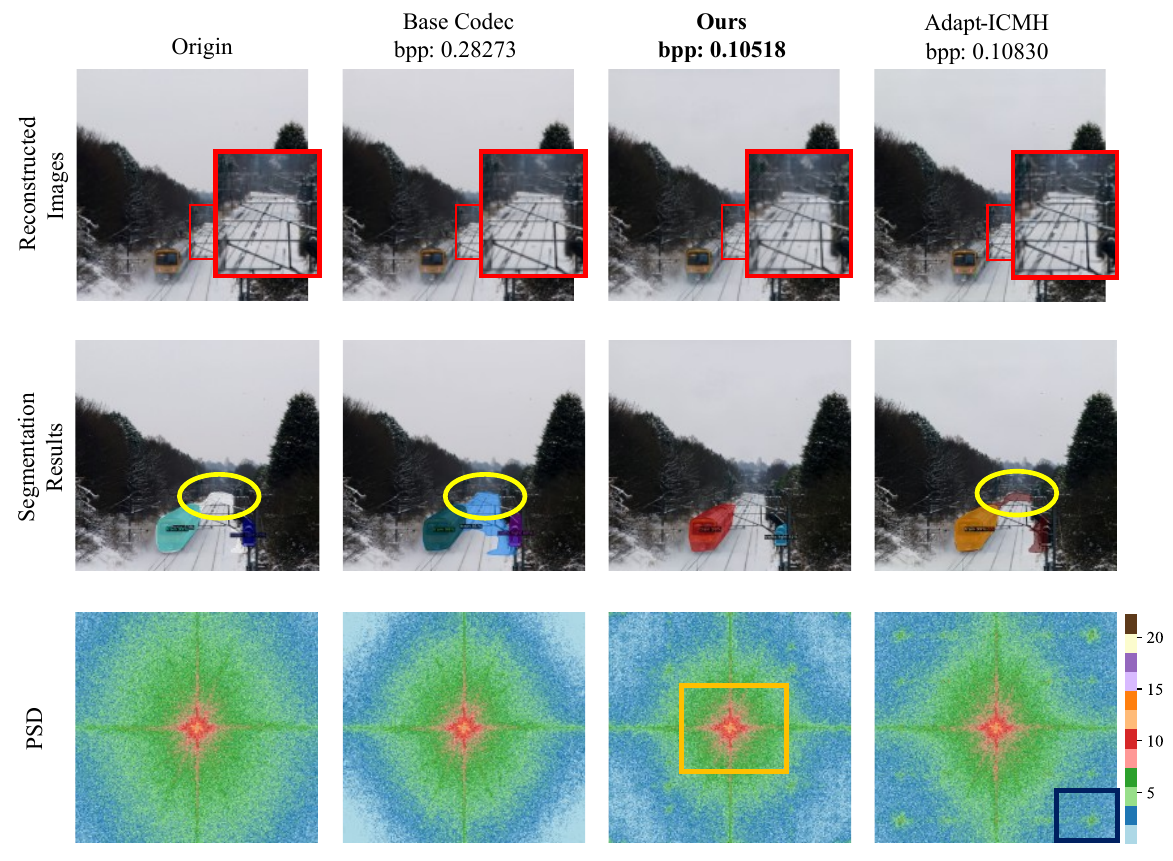} 
	\caption{More segmentation qualitative results.}
	\label{42563}
\end{figure*}

\begin{figure*}[!htbp]
	\centering
	\includegraphics[width=1.8\columnwidth, height=10.2cm]{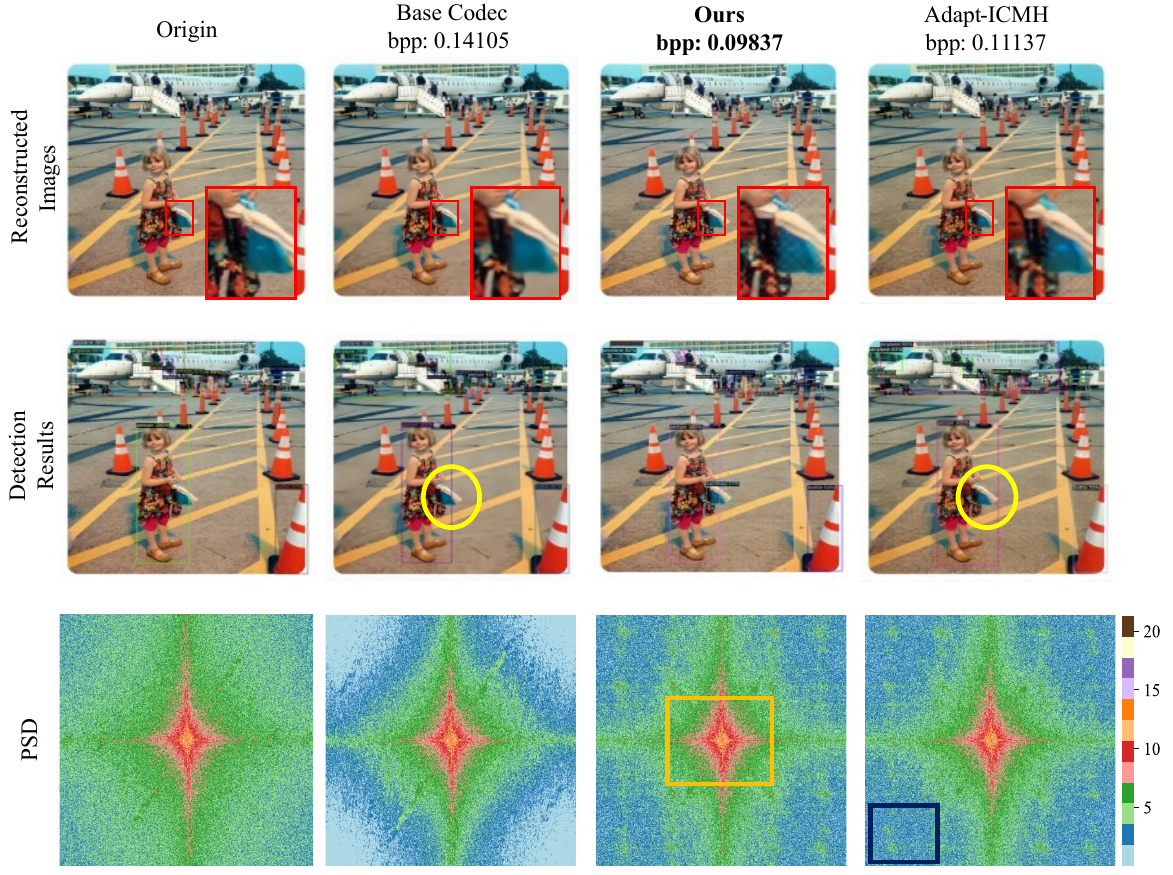} 
	\caption{More detection qualitative results.}
	\label{381639}
\end{figure*}

\begin{figure*}[!htbp]
	\centering
	\includegraphics[width=1.8\columnwidth, height=10.2cm]{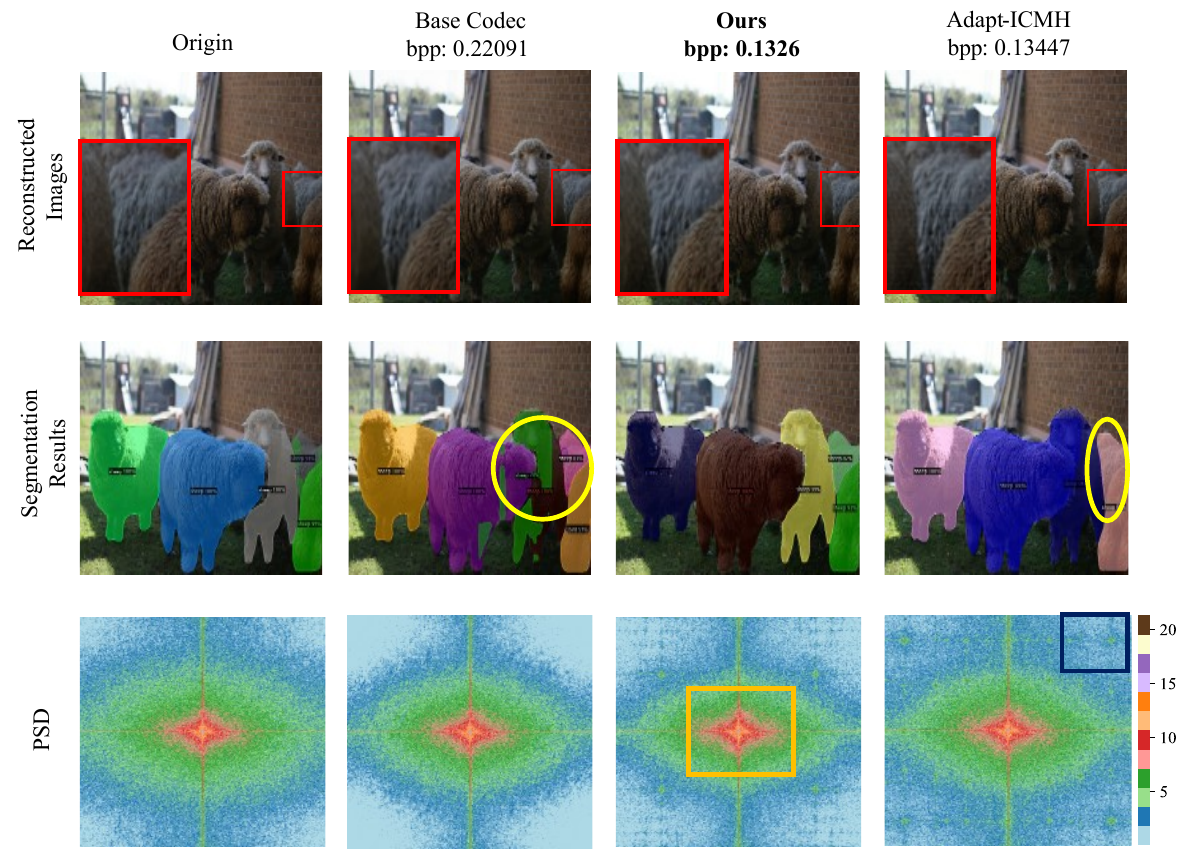} 
	\caption{More segmentation qualitative results.}
	\label{5992}
\end{figure*}

To complement the visual comparisons presented in the main paper, we provide additional qualitative results for the S$^2$-CoT framework in this section. As shown in the \cref{254368} - \cref{5992}, our method consistently achieves superior visual fidelity, demonstrating enhanced structural integrity and fewer spurious high-frequencies compared to competing methods, even at low bitrates. These results further substantiate the claim that S$^2$-CoT effectively preserves the critical semantic information necessary for both human perception and downstream machine vision tasks.

\section{Generalization of Synergy to Advanced Entropy Models Across Diverse Base Codecs}
\label{sec:supgeneralizationSynergy}

While \cref{sec:supTheoreticalAnalysis} establishes the theoretical foundation using a standard Gaussian hyperprior model, state-of-the-art learned image compression frameworks employ significantly more sophisticated entropy modeling mechanisms to capture complex dependencies. In this section, we mathematically demonstrate that the proposed S$^2$-CoT paradigm is not limited to basic architectures but establishes a unified synergistic mechanism across diverse entropy models. 

By mathematically decomposing the probability estimation processes of \textit{Cheng2020-anchor} (Autoregressive GMM), \textit{DCAE} (Dictionary-based Prior) and \textit{ELIC} (Spatial-Channel Context), we show that the hyperprior feature $\boldsymbol{\psi}$ serves as the fundamental anchor for probability parameterization in all cases. Consequently, our SCA, by refining this shared context, effectively propagates synergistic corrections through these complex probability chains.

\subsection{Synergy in Autoregressive GMMs}
\label{subsec:supSynergyInAutoregressiveGMMs}

Sophisticated codecs like $ \textit{Cheng2020-anchor} $ \cite{Cheng2020} enhance coding efficiency by combining channel-wise autoregression with Gaussian Mixture Models (GMM). The conditional probability for a latent $\hat{y}_i$ is formulated as:
\begin{equation}
	p_{\hat{y}_i | \hat{\mathbf{z}}, \hat{\mathbf{y}}_{<i}}(\hat{y}_i | \cdot) = \sum_{k=1}^K w_i^{(k)} \mathcal{N} \left(\mu_i^{(k)}, {\sigma_i^{(k)}}^2 \right) * \mathcal{U}\left( -\frac{1}{2}, \frac{1}{2}\right),
\end{equation}
where the mixture parameters are jointly predicted by an entropy parameter network $g_{ep}$:
\begin{equation}
	\left(w_i^{(k)}, \mu_i^{(k)}, \sigma_i^{(k)}\right)_{k=1}^K = g_{ep}(\boldsymbol{\psi}, \boldsymbol{\phi}_i; \boldsymbol{\theta}_{ep}).
\end{equation}
Here, $\boldsymbol{\phi}_i$ represents the autoregressive context from previous latents $\hat{y}_{<i}$, and $\boldsymbol{\psi} = h_s(\hat{\mathbf{z}}; \boldsymbol{\theta}_{h_s})$ is the hyperprior feature derived from the hyper-decoder.

\textbf{Synergy Mechanism:} Even with the introduction of temporal context $\boldsymbol{\phi}_i$, the GMM parameters remain functionally dependent on the hyperprior feature $\boldsymbol{\psi}$. By inserting the SCA into the hyperprior pathway, our method dynamically calibrates $\boldsymbol{\psi}$ to $\boldsymbol{\psi}' = \text{SCA}(\boldsymbol{\psi})$. This calibrated feature acts as a corrected condition for $g_{ep}$, guiding the GMM to generate an optimal mixture distribution that aligns with the SFA-adapted latent space, thereby minimizing the coding cost $\mathbb{E}[-\log_2 p]$.

\subsection{Synergy in Dictionary Entropy Models}
\label{subsec:supSynergyInDictionaryEntropyModels}

The state-of-the-art $ \textit{DCAE} $ codec \cite{Lu2025} introduces a dictionary-based cross-attention mechanism to capture long-range dependencies. Its bitrate calculation involves a dictionary-based prior $\boldsymbol{\delta}$:
\begin{equation}
	R(\hat{\mathbf{y}}) = \sum_{k} \mathbb{E} \left[ -\log_2 p \left( \hat{y}^k \mid \boldsymbol{\psi}, \boldsymbol{\phi}^k, \boldsymbol{\delta}^k; \boldsymbol{\theta}_{ep} \right) \right],
\end{equation}
where the dictionary prior is computed via cross-attention: $\boldsymbol{\delta}^k = \text{Attn}([\boldsymbol{\psi}, \boldsymbol{\phi}^k], \mathbf{D})$.

\textbf{Synergy Mechanism:} In this architecture, the hyperprior $\boldsymbol{\psi}$ serves a dual role: it is a direct condition for the entropy parameters and a query/key component for querying the learned dictionary $\mathbf{D}$. Therefore, the distributional shift analyzed in \cref{sec:supTheoreticalAnalysis} would be exacerbated here, as an uncalibrated $\boldsymbol{\psi}$ would retrieve erroneous priors from the dictionary. Our S$^2$-CoT framework resolves this by tuning $\boldsymbol{\psi}$ via SCA, ensuring that the dictionary lookup retrieves task-relevant priors. This validates that SCA is not merely a local fix but a fundamental component that restores the operational integrity of advanced attention-based models.

\subsection{Context Synergy in Spatial-Channel}
\label{subsec:supContextSynergyInSpatial-Channel}

Recent models like  $ \textit{ELIC} $ codec \cite{He2022a} further decouple the latent space into anchor ($y_{ac}$) and non-anchor ($y_{na}$) groups to exploit uneven spatial-channel correlations. The total bitrate is decomposed as:
\begin{align}
	R(\hat{\mathbf{y}}) &= \sum \left( R_{\hat{y}_{ac}} + R_{\hat{y}_{na}} \right), \\
	R_{\hat{y}_{ac}} &= \mathbb{E} \left[ -\log_2 p \left( \hat{y}_{ac} \mid \boldsymbol{\psi}, \boldsymbol{\theta}_{ch} \right) \right], \\
	R_{\hat{y}_{na}} &= \mathbb{E} \left[ -\log_2 p \left( \hat{y}_{na} \mid \boldsymbol{\psi}, \boldsymbol{\theta}_{ch}, \boldsymbol{\theta}_{sp} \right) \right].
\end{align}
Here, $\boldsymbol{\theta}_{ch}$ and $\boldsymbol{\theta}_{sp}$ denote the parameters for channel and spatial context transforms, respectively. 

\textbf{Synergy Mechanism:} Crucially, both the anchor and non-anchor probability estimates are conditioned on the shared hyperprior context $\boldsymbol{\psi}$. The SCA essentially functions as a global context modulator. By refining $\boldsymbol{\psi}$, the SCA simultaneously optimizes the base probability for anchor latents and the conditional priors for non-anchor latents. This ensures that the complex spatial-channel grouping logic in $ \textit{ELIC} $ operates on statistically aligned features, preserving the synergy between the structural transform and the contextual entropy model.

\subsection{Unified Perspective}
\label{subsec:supUnifiedPerspective}

Across all the aforementioned formulations, spanning from simple Gaussian models to complex dictionary-augmented GMMs, a pivotal common structural invariant emerges: the hyperprior pathway provides the core foundational latent-side information $\boldsymbol{\psi}$ that parameterizes the underlying conditional distributions.

Our S$^2$-CoT framework leverages this invariant. Instead of redesigning adapters for every specific entropy sub-module (\eg, autoregressive heads or attention blocks), we strategically place the SCA to refine the root information $\boldsymbol{\psi}$. This establishes a model-agnostic synergy, where a lightweight modulation at the semantic root effectively propagates statistical alignment throughout the entire entropy modeling hierarchy.

\clearpage
\newpage

%
%\clearpage
%\newpage
%{
%	\small
%	\bibliographystyle{ieeenat_fullname}
%	\bibliography{all_right}
%}